\documentclass{article}

\usepackage{microtype}
\usepackage{graphicx}
\usepackage{subfigure}
\usepackage{booktabs} 

\usepackage{hyperref}

\usepackage[preprint]{style_file}

\usepackage{amsmath}
\usepackage{amssymb}
\usepackage{mathtools}
\usepackage{amsthm}

\usepackage[capitalize,noabbrev]{cleveref}

\usepackage{stmaryrd}

\usepackage{algorithmicx}
\usepackage{algpseudocode}

\theoremstyle{plain}
\newtheorem{theorem}{Theorem}[section]

\newtheorem{lemma}[theorem]{Lemma}

\theoremstyle{definition}

\theoremstyle{remark}

\usepackage[textsize=tiny]{todonotes}

\icmltitlerunning{Fairness with Exponential Weights}

\begin{document}

\twocolumn[
\icmltitle{Fairness with Exponential Weights}



\icmlsetsymbol{equal}{*}

\begin{icmlauthorlist}
\icmlauthor{Stephen Pasteris}{ati}
\icmlauthor{Chris Hicks}{ati}
\icmlauthor{Vasilios Mavroudis}{ati}
\end{icmlauthorlist}

\icmlaffiliation{ati}{The Alan Turing Institute, London, UK}

\icmlcorrespondingauthor{Stephen Pasteris}{spasteris@turing.ac.uk}
\icmlcorrespondingauthor{Chris Hicks}{c.hicks@turing.ac.uk}
\icmlcorrespondingauthor{Vasilios Mavroudis}{vmavroudis@turing.ac.uk}

\icmlkeywords{Bandits, Online Learning, Exponential Weights, Group Fairness, Statistical Parity}

\vskip 0.3in
]



\printAffiliationsAndNotice{}  

\newcommand{\nc}[1]{\newcommand{#1}}
\nc{\bs}[1]{\boldsymbol{#1}}
\nc{\be}{\begin{equation*}}
\nc{\ee}{\end{equation*}}
\nc{\indi}[1]{\llbracket#1\rrbracket}
\nc{\emp}[1]{\emph{#1}}
\nc{\cons}{\mathcal{X}}
\nc{\acs}{\mathcal{A}}
\nc{\pd}{\mu}
\nc{\arp}{\tilde{\pi}}
\nc{\arc}{x}
\nc{\ac}{a}
\nc{\blank}{\circ}
\nc{\ntr}{T}
\nc{\cont}[1]{x_{#1}}
\nc{\lost}[1]{\ell_{#1}}
\nc{\loc}[2]{\ell_{#1,#2}}
\nc{\act}[1]{a_{#1}}
\nc{\pols}{\mathcal{P}}
\nc{\polt}[1]{\pi_{#1}}
\nc{\draw}{\sim}
\nc{\reg}[1]{R(#1)}
\nc{\bexpt}[1]{\mathbb{E}\left[#1\right]}
\nc{\seq}[2]{\langle#1\,|\,#2\rangle}
\nc{\pow}{\mathcal{U}}
\nc{\pob}{\mathcal{V}}
\nc{\apd}{\mu}
\nc{\cpdd}[2]{\mathbb{P}_{#1}[#2]}
\nc{\cedd}[3]{\mathbb{E}_{#1}[#2\,|\,#3]}
\nc{\mpd}{\tilde{\mu}}
\nc{\alos}{\bs{\ell}}
\nc{\arcp}{y}
\nc{\edt}[1]{{\mu}_{#1}}
\nc{\alg}{\textsc{Few}}
\nc{\nmc}{N}
\nc{\nma}{K}
\nc{\splx}[1]{\Delta_{#1}}
\nc{\loft}[1]{\ell_{#1}}
\nc{\nat}{\mathbb{N}}
\nc{\scd}{\mu}
\nc{\nmg}{M}
\nc{\prg}[1]{\mathcal{G}_{#1}}
\nc{\pri}{i}
\nc{\arbs}{\mathcal{B}}
\nc{\arbe}{z}
\nc{\sps}[1]{\mathcal{F}(#1)}
\nc{\arf}{f}
\nc{\cpol}[1]{\tilde{\pi}_{#1}}
\nc{\ccp}{\tilde{\pi}}
\nc{\efs}{\mathcal{E}(\bmu)}
\nc{\bpol}{\bs{\pi}}
\nc{\apol}{\tilde{\pi}}
\nc{\sreg}[1]{R(#1)}
\nc{\dreg}[1]{R(#1)}
\nc{\cbp}{\tilde{\bs{\pi}}}
\nc{\pthl}[1]{\Lambda(#1)}
\nc{\mmex}{\omega}
\nc{\arbn}{Z}
\nc{\losd}{\lambda}
\nc{\pred}{P}
\nc{\val}{V}
\nc{\cednc}[2]{\mathbb{E}_{#1}[#2]}
\nc{\gmd}[1]{m_{#1}}
\nc{\prj}{j}
\nc{\prb}[1]{\mathbb{P}[#1]}
\nc{\cprb}[2]{\mathbb{P}[#1\,|\,#2]}
\nc{\cdn}{z}
\nc{\dist}{D}
\nc{\spt}[1]{\edt{#1}}
\nc{\rpol}[1]{\xi_{#1}}
\nc{\cst}[1]{\mathcal{C}(\edt{#1})}
\nc{\set}[2]{\{#1\,|\,#2\}}
\nc{\pig}[1]{k_{#1}}
\nc{\cop}[1]{\mathcal{Z}_t}
\nc{\cpv}{\phi}
\nc{\sed}{\sigma}
\nc{\ppo}[1]{\beta_{#1}}
\nc{\der}{\nabla}
\nc{\lm}[2]{\lambda_{t,#1}(#2)}
\nc{\ppt}[1]{\beta^\dag_{#1}}
\nc{\clp}[1]{\tilde{\beta}_{#1}}
\nc{\crf}[1]{\mu_{#1}}
\nc{\csf}[2]{h_{#1}(#2)}
\nc{\vpo}[1]{\psi_{#1}}
\nc{\mih}[1]{h^*(#1)}
\nc{\dif}[2]{\delta_{#1}(#2)}
\nc{\difs}[1]{\delta'_{#1}}
\nc{\slf}[1]{y_{#1}}
\nc{\expt}[1]{\mathbb{E}[#1]}
\nc{\cnd}[1]{\gamma_{#1}}
\nc{\real}{\mathbb{R}}
\nc{\edraw}[3]{\mathbb{E}_{#1\draw#2}[#3]}
\nc{\gra}[1]{\lambda_{#1}}
\nc{\brd}{B}
\nc{\mdp}[1]{\xi'_{#1}}
\nc{\lr}{\hat{\eta}}
\nc{\pra}{\eta}
\nc{\arpp}{\hat{\pi}}
\nc{\ent}{F}
\nc{\egr}[1]{\xi_{#1}}
\nc{\eps}{\mathcal{S}}
\nc{\gam}[2]{\tilde{\omega}_{#1,#2}}
\nc{\gaa}[1]{\tilde{\omega}_{#1}}
\nc{\sqb}[1]{#1}
\nc{\edi}[1]{\tilde{d}_{#1}}
\nc{\edit}[1]{d^\dag_{#1}}
\nc{\ugm}[1]{\xi_{#1}}
\nc{\lgm}[1]{\tilde{\xi}_{#1}}
\nc{\ed}[2]{\tilde{\delta}_{#1,#2}}
\nc{\inr}[1]{r_{#1}}
\nc{\hconst}{\hat{c}}
\nc{\const}{c}
\nc{\mac}{L}
\nc{\imp}[1]{\psi_{t}}
\nc{\unc}[1]{h_{#1}}
\nc{\prit}{j}
\nc{\cdi}[2]{\delta_{#1,#2}}
\nc{\cer}[1]{\tilde{\beta}_{#1}}
\nc{\cra}[1]{\beta_{#1}}
\nc{\la}{:=}
\nc{\osa}[1]{\tilde{a}_{#1}}
\nc{\kal}[1]{\kappa_{#1}}
\nc{\kau}[1]{\kappa_{#1}'}
\nc{\epl}[1]{\mathcal{Z}_{#1}}
\nc{\spr}{\gamma}
\nc{\arfu}{f}
\nc{\hli}{\noindent \rule{5cm}{0.4pt} \vspace{4pt}}
\nc{\vsp}{\vspace{8pt}}
\nc{\ntgv}{\bs{\tau}}
\nc{\ntg}[1]{\tau^\dag_{#1}}
\nc{\uir}{\rho^\dag}
\nc{\frp}{\xi^\dag}
\nc{\mnu}{v^\dag}
\nc{\anu}{b^\dag}
\nc{\egm}[1]{\omega^\dag_{#1}}
\nc{\qma}[1]{q_{#1}}
\nc{\qmi}[1]{q'_{#1}}
\nc{\pat}[1]{i_{#1}}
\nc{\lla}{\leftarrow}
\nc{\ntgt}[2]{\tau_{#1,#2}}
\nc{\cdag}{\mathcal{C}^\dag}
\nc{\psih}[1]{\psi'_{#1}}
\nc{\pih}[1]{\pi'_{#1}}
\nc{\hsu}[1]{h_{#1}}
\nc{\nrml}[1]{\sigma_{#1}}
\nc{\kapu}[1]{\tilde{\kappa}_{#1}}
\nc{\kapl}[1]{\tilde{\kappa}'_{#1}}
\nc{\kanu}[1]{\kappa_{#1}}
\nc{\kanl}[1]{\kappa'_{#1}}
\nc{\rem}[1]{h_{#1}}
\nc{\tgr}[1]{g_{#1}}
\nc{\cpvd}{\phi^*}
\nc{\pritt}{k}
\nc{\exptc}[2]{\mathbb{E}[#1\,|\,#2]}
\nc{\bexptc}[2]{\mathbb{E}\left[#1\,\Bigg|\,#2\right]}
\nc{\probc}[2]{\mathbb{P}[#1\,|\,#2]}
\nc{\midl}[1]{\lambda'_{#1}}
\nc{\bmu}{\bs{\mu}}
\nc{\cond}[2]{#1^{|#2}}
\nc{\tpdt}[1]{\tilde{\mu}_{#1}}
\nc{\tpdb}{\tilde{\bs{\mu}}}
\nc{\gps}{\mathcal{C}}
\nc{\acp}{b}
\nc{\bx}{\bs{x}}
\nc{\inp}{\pi}
\nc{\rplus}{\real_+}
\nc{\spg}{\mathcal{G}}
\nc{\spgp}{\mathcal{H}}
\nc{\losp}[1]{\ell'_{#1}}
\nc{\rele}{\tilde{B}}
\nc{\ublo}[1]{\hat{\ell}_{#1}}
\nc{\ucu}[1]{\zeta_{#1}}
\nc{\brs}{B^*}
\nc{\algt}{\textsc{FexEx2}}
\nc{\cpd}{\mu}
\nc{\exs}{\mathcal{H}}
\nc{\iwf}{\vartheta}
\nc{\ex}{e}
\nc{\aru}{u}
\nc{\iwfp}{\vartheta'}
\nc{\upds}[3]{\textsc{Update}(#1,#2,#3)}
\nc{\qrys}{\textsc{Query}}
\nc{\trp}{\varphi}
\nc{\iwfh}{\vartheta^*}
\nc{\iwt}[1]{\vartheta_{#1}}
\nc{\cuts}[1]{\chi(#1)}
\nc{\comp}{\tilde{\pi}}
\nc{\jupds}{\textsc{Update}}
\nc{\jqrys}{\textsc{Query}}
\nc{\gad}{\beta}
\nc{\rwcs}{\Phi}
\nc{\bcpd}{p}
\nc{\arl}{\ell}
\nc{\scr}{\tilde{R}}
\nc{\genr}{R^\dag}
\nc{\str}{R}
\nc{\ups}{\textsc{Update}}
\nc{\qus}{\textsc{Query}}
\nc{\opd}{\pi'}
\nc{\aws}{\mathcal{W}}
\nc{\spnm}{z}
\nc{\spnp}{z'}
\nc{\sfw}{\vartheta^\dag}
\nc{\awm}{w}
\nc{\vpd}{\varphi}
\nc{\res}[2]{[#1|#2]}
\nc{\iwp}[2]{\vartheta_{#1,#2}}
\nc{\sre}{\tilde{B}}
\nc{\geft}[1]{\gef_{#1}}
\nc{\qsc}[1]{\textsc{Query}[#1]}
\nc{\usc}[1]{\textsc{Update}[#1]}
\nc{\gamn}[1]{\tilde{\omega}_{#1}}
\nc{\edn}[1]{\tilde{\delta}_{#1}}
\nc{\cdin}[1]{\delta_{#1}}
\nc{\iwtp}[1]{\vartheta'_{#1}}
\nc{\re}{B}
\nc{\avt}{\beta}
\nc{\gef}{\nu}
\nc{\prcs}{\mathcal{C}}
\nc{\prc}{c}
\nc{\prct}[1]{c_{#1}}
\nc{\cds}{\mathcal{D}}
\nc{\ins}{\mathcal{I}}
\nc{\ibs}{\splx{\exs}}
\nc{\acd}{\varphi}
\nc{\acdt}[1]{\varphi_{#1}}
\nc{\expr}{\mathcal{E}}
\nc{\supp}[1]{\mathcal{G}_{#1}}
\nc{\ics}{\mathcal{X}'}
\nc{\vct}[1]{x'_{#1}}
\nc{\ftr}[1]{\mathcal{X}^*_{#1}}
\nc{\tmd}[2]{\tilde{\mu}_{#1,#2}}
\nc{\ctr}[2]{\mathcal{X}_{#1,#2}}
\nc{\apm}{\mu'}
\nc{\cdp}{h}
\nc{\nsm}{n}
\nc{\asc}[1]{\mathcal{U}(#1)}
\nc{\empr}{p}
\nc{\empa}{p^*}
\nc{\aam}{\mu^*}
\nc{\ass}{\mathcal{S}}
\nc{\lch}[1]{{\triangleleft}(#1)}
\nc{\rch}[1]{{\triangleright}(#1)}
\nc{\hpv}{\delta}
\nc{\eap}{\epsilon}
\nc{\lefs}[2]{\mathcal{L}_{#1,#2}}
\nc{\des}[2]{\mathcal{D}_{#1}(#2)}
\nc{\pp}[1]{\pi'_{#1}}
\nc{\ps}[1]{\pi^\dag_{#1}}
\nc{\vhe}[2]{\nu_{#1}(#2)}
\nc{\bap}{\beta}
\nc{\bapp}{\beta'}
\nc{\baph}{\hat{\beta}}
\nc{\tpol}[1]{\hat{\pi}_{#1}}
\nc{\bapt}[2]{\tilde{\beta}_{#1,#2}}
\nc{\eaph}{\hat{\eap}}
\nc{\prob}[2]{\mathbb{P}[#1\,|\,#2]}
\nc{\trele}{\tilde{B}}
\nc{\vtpd}{\hat{\vartheta}}
\nc{\uas}{\mathcal{B}}
\nc{\pnt}[1]{{\uparrow}(#1)}
\nc{\pst}{p^*}
\nc{\psbf}[1]{\mathcal{X}^\dag(#1)}
\nc{\vdp}{d}
\nc{\den}{\delta'}
\nc{\gan}{\gamma}
\nc{\ndes}[1]{\mathcal{L}_{#1}}
\nc{\bdf}[1]{\beta'_{#1}}
\nc{\nvhe}[1]{h'_{#1}}
\nc{\omwt}[1]{\omega_{#1}}
\nc{\prwcs}[1]{\theta(#1)}

\begin{abstract}
Motivated by the need to remove discrimination in certain applications, we develop a meta-algorithm that can convert any efficient implementation of an instance of Hedge (or equivalently, an algorithm for discrete bayesian inference) into an efficient algorithm for the equivalent contextual bandit problem which guarantees exact statistical parity on every trial. Relative to any comparator with statistical parity, the resulting algorithm has the same asymptotic regret bound as running the corresponding instance of Exp4 for each protected characteristic independently. Given that our Hedge instance admits non-stationarity we can handle a varying distribution with which to enforce statistical parity with respect to, which is useful when the true population is unknown and needs to be estimated from the data received so far. Via online-to-batch conversion we can handle the equivalent batch classification problem with exact statistical parity, giving us results that we believe are novel and important in their own right.
\end{abstract}

\section{Introduction}
Often, an important ethical requirement of a classifier is to be unbiased with respect to certain \emp{protected characteristics} such as ethnicity, gender, age or disability. This is formalised by the notion of \emp{statistical parity} which is defined relative to a probability distribution $\rho$ over the instance space $\ins=\prcs\times\cons$, where $\prcs$ is the set of \emp{protected characteristics} which we have to be unbiased with respect to, and any $\arc\in\cons$ (called a \emp{context}) represents a collection of other, unprotected characteristics that allow us to learn. A stochastic classifier (a.k.a. \emp{policy}) of the instance space $\ins$ is said to have \emp{statistical parity} with respect to $\rho$ if and only if, when an instance $(\prc,\arc)$ is drawn from $\rho$ and then the policy stochastically selects a corresponding class (a.k.a. \emp{action}) $\ac$, then: 
\be
\mathbb{P}[\ac=\acp\,|\,\prc=d]=\mathbb{P}[\ac=\acp\,|\,\prc=d']
\ee
for all actions $\acp$ and all protected characteristics $d,d'\in\prcs$. In other words, the policy treats, overall, each protected group (i.e. instances sharing a particular protected characteristic) identically.

In this paper we consider the task of enforcing statistical parity in both online and batch tasks. In the online task we play in a sequence of trials, on each trial receiving an instance and then choosing an action. At the end of any trial we view either the loss (or equivalently, a reward) associated with each action (the \emp{full-information} setting) or the loss associated with just the action we select (the \emp{bandit} setting). We consider the most general \emp{fully-adversarial} problem where there are no assumptions on the generation of the instance/loss sequence, but we believe our results are also novel for the i.i.d. stochastic special case. 

Perhaps the most fundamental and widely-applicable algorithm for the full-information setting with no protected characteristics is \textsc{Hedge} \cite{Freund1997ADG}, which maintains a probability distribution over all possible mappings from contexts to actions (a.k.a. \emp{experts}) and can be applied to different problems via the choice of the initial probability distribution (a.k.a. the \emp{inductive bias}). It is often possible, when given a particular inductive bias, to implement \textsc{Hedge} extremely efficiently. Specifically, any algorithm for discrete bayesian inference with a prior matching the inductive bias is also an algorithm for implementing \textsc{Hedge}. The work \cite{Auer2002TheNM} gave a meta-algorithm \textsc{Exp4} for converting any efficient implementation of \textsc{Hedge} into an efficient algorithm for the bandit setting. In this paper we give a meta-algorithm \alg\ ({\bf F}airness with {\bf E}xponential {\bf W}eights) for converting any efficient implementation of \textsc{Hedge}, which we call the \emp{base algorithm}, into an efficient algorithm for the bandit setting that guarantees statistical parity exactly on every trial. The resulting algorithm can then also be used for batch classification under statistical parity constraints via \emp{online-to-batch conversion} \cite{CesaBianchi2001OnTG}.

\alg\ has excellent performance guarantees, both in terms of the quality of its selections and its computational complexity. Relative to any comparator policy (or policy sequence) with statistical parity, \alg\ has exactly the same asymptotic regret bound as running the equivalent instance of \textsc{Exp4} for each protected characteristic independently. Given an efficient-enough base algorithm, \alg\ has a per-trial time complexity of only $\mathcal{O}(\nmg\nmc\nma)$ where $\nmg$, $\nmc$ and $\nma$ are the number of protected characteristics, number of contexts and number of actions respectively. Note that this is linear in the dimensionality of the policy space. We will also show how to use an algorithm of \cite{Pasteris2023NearestNW} as our base algorithm in order to efficiently handle massive and potentially infinite context spaces that can be hierarchically decomposed (such as euclidean space), achieving a per-trial time complexity of $\mathcal{O}(\ntr\nma)$ where $\ntr$ is the number of trials. The space complexity is always equal to that of maintaining an instance of the base algorithm for each protected characteristic. These performance guarantees transfer over to batch classification problems. We note that, due to the equivalence of \textsc{Hedge} and bayesian inference, if no efficient implementation exists for our inductive bias, we could use an algorithm for approximate bayesian inference instead (and still obtain exact statistical parity).

Sometimes we will not know the population (i.e. the distribution with which to maintain statistical parity with respect to) a-priori, and hence we will need to use the empirical distribution of instances seen so far to approximate it. In order to do this we note that \alg\ has the ability to maintain statistical parity with respect to a dynamic distribution as long as a policy exists which always has statistical parity. With this result under our belt we give the following two examples of learning the distribution. First, we expand the context space to all context/trial pairs so that we can use \textsc{FixedShare} \cite{Herbster1995TrackingTB} as our base algorithm, which allows us to always maintain statistical parity with respect to the empirical distribution of instances seen so far, as well as adapting to changing environments. However, this does not exploit an inductive bias on the contexts and, for the empirical distribution to be meaningful, specific contexts need to be seen multiple times (although this algorithm can handle massive/infinite context spaces when given a clustering a-priori). Second, when given a (potentially massive) context space that can be hierarchically decomposed, and when the instance sequence is drawn i.i.d. from the population, we show how to maintain approximate statistical parity (on every trial) with high probability, again using an algorithm of \cite{Pasteris2023NearestNW} as our base algorithm.

The methodology of \alg\ should also be applicable to certain other exponentiated gradient algorithms (or even other types of gradient-based algorithms) which aren't actually implementations of \textsc{Hedge}. For example, we should be able to apply the methodology to the \textsc{CBA} algorithm of \cite{Pasteris2024BanditsWA} which would allow us to utilise confidence-rated expert advice, and attain true reward bounds, when we have an \emp{abstention} action/class and a finite context set.

We now discuss the failure of trying to add the statistical parity constraints directly to \textsc{Exp4}. \textsc{Exp4} is a modification of \textsc{Hedge} which implements mirror descent over the probability simplex of the experts. Hence, the statistical parity constraints can be added directly to the mirror descent. However, this would require a relative entropy projection onto the constraint set. To do this we construct the lagrangian and set its derivative to zero. Writing the primal variables in terms of the Lagrange multipliers is staightforward but, as far as we are aware, finding the correct values of the Lagrange multipliers is analytically intractable. Since the projection is a convex optimisation problem we could use methods to approximate it instead. However, in addition to only giving an approximation these methods would, as far as we are aware, take exponential time due to the exponential dimensionality. In the special case in which our base algorithm treats each context independently (i.e. there is effectively no inductive bias) the problem reduces to one with polynomial dimension, but as far as we are aware all approximation methods would still be very inefficient compared to \alg\ here.

There has been much work on designing efficient algorithms for online convex optimisation which (unlike \alg) violate convex constraints but whose cumulative constraint violation can be bounded. In particular, the work \cite{Castiglioni2022AUF} can handle the bandit problem and has a cumulative constraint violation of $\mathcal{O}(\ntr^{3/4})$ (where the $\mathcal{O}$ suppresses other values not dependent on $\ntr$) but at the cost of an additional multiplicative factor $\mathcal{O}(\nmg^{1/2}\ntr^{1/4})$ to the regret. The work \cite{Yu2020ALC} managed to achieve a cumulative constraint violation of $\mathcal{O}(\ntr^{1/2})$ but fails to work in our setting, being incompatible with the bandit problem and having a polynomial dependence on the dimensionality (which in our case is exponential) in the constraint violation, regret and time complexity.

We will discuss further related works in Appendix \ref{arwsec}.

\section{Notation}
In this section we introduce the notation used in this paper. We define $\rplus$ to be the set of non-negative real numbers. We define $\nat$ to be the set of natural numbers excluding $0$. Given $\arbe\in\nat$ we define:
\be
[\arbe]:=\{\arbe'\in\nat\,|\,\arbe'\leq\arbe\}
\ee
Given some finite set $\arbs$ we define $\splx{\arbs}$ to be the set of probability distributions over $\arbs$. That is, $\splx{\arbs}$ is the set of all $\rho:\arbs\rightarrow[0,1]$ with:
\be
\sum_{\arbe\in\arbs}\rho(\arbe)=1
\ee
We call the set $\splx{\arbs}$ a \emp{simplex}. Given sets $\arbs$ and $\arbs'$ and a function $\arf$ with domain $\arbs\times\arbs'$ we define, for all $\arbe\in\arbs$, the function $\arf(\arbe,\blank)$ to be the function that maps each $\arbe'\in\arbs'$ to $\arf(\arbe,\arbe')$. Given a predicate $P$ we define $\indi{P}:=1$ if $P$ is true and $\indi{P}:=0$ otherwise. In our problem definitions we will refer to ourselves as \emp{Learner} and refer to the environment that we interact with as \emp{Nature}.

\section{Fundamental Concepts}

Before we describe our results we will, in this section, introduce the required fundamental concepts.

\subsection{Statistical Parity}

Here we describe what it means for a classifier to satisfy \emp{statistical parity}. In this paper we will use the following terminology for classification: a \emp{policy} is a stochastic classifier and an \emp{action} is a class.

We have finite sets $\prcs$, $\cons$ and $\acs$ of \emp{protected characteristics}, \emp{contexts} and \emp{actions} respectively. We define:
\be
\nmg:=|\prcs|~~~~~~;~~~~~\nmc:=|\cons|~~~~~~;~~~~~\nma:=|\acs|
\ee
We define the set of \emp{instances} as:
\be
\ins:=\prcs\times\cons
\ee
Note that an instance is a context with an associated protected characteristic. The idea of statistical parity is that, when given an instance, we want to stochastically generate an action that is unbiased with respect to the protected characteristic.

To formalise this notion we define the space of \emp{targets} to be the set:
\be
\cds:=\left\{\apd\in[0,1]^{\prcs\times\cons}\,|\,\forall \prc\in\prcs,\, \apd(\prc,\blank)\in\splx{\cons}\right\}
\ee
Note that any target $\apd\in\cds$ represents a probability distribution over $\ins$ such that for any instance $(\prc,\arc)\in\ins$ we have that $\apd(\prc,\arc)$ is the probability of the instance $(\prc,\arc)$ conditioned on the protected characteristic being $\prc$.

We define the space of \emp{policies} to be the set:
\be
\pols:=\left\{\inp\in[0,1]^{\prcs\times\cons\times\acs}\,|\,\forall (\prc,\arc)\in\ins\,,\,\inp(\prc,\arc,\blank)\in\splx{\acs}\right\}
\ee
Note that when a policy $\inp\in\pols$ receives an instance $(\prc,\arc)\in\ins$ it draws an action from the probability distribution $\inp(\prc,\arc,\blank)$.

For any $\apd\in\cds$ we define $\sps{\apd}$ to be the set of all $\inp\in\pols$ in which for all $\prc,\prc'\in\prcs$ and $\ac\in\acs$ we have:
\be
\sum_{\arc\in\cons}\apd(\prc,\arc)\inp(\prc,\arc,\ac)=\sum_{\arc\in\cons}\apd(\prc',\arc)\inp(\prc',\arc,\ac)
\ee
We say that $\sps{\apd}$ is the set of policies that have \emp{statistical parity} with respect to $\apd$. Note that a policy $\inp\in\pols$ has statistical parity with respect to $\apd$ iff for all protected characteristics $\prc\in\prcs$ and actions $\ac\in\acs$, when $\arc$ is drawn from $\apd(\prc,\blank)$ and then $\ac'$ is drawn from $\inp(\prc,\arc,\blank)$, the probability that $\ac'=\ac$ is independent of $\prc$. In other words, $\inp$ is unbiased with respect to the protected characteristic. When it is clear what $\apd$ is we will call such a policy \emp{fair}.

\subsection{Experts and the Inductive Bias}

Unless the context set is relatively small, we will require an \emp{inductive bias} to learn well. Such an inductive bias is given by the problem at hand. In this paper we define an inductive bias to be any distribution:
\be
\iwf\in\splx{\acs^\cons}
\ee
For notational simplicity we define:
\be
\exs:=\acs^\cons
\ee
which is the \emp{hypothesis space}. We will refer to the elements of $\exs$ as \emp{experts}, so that each expert is a mapping from $\cons$ into $\acs$ and is hence a deterministic classifier of the contexts. If the inductive bias gives a higher probability to some expert then it means that we are more confident that that expert is likely a good classifier.

In fact, there is nothing to stop us from adding stochastic experts to the hypothesis space, but in this paper we keep them deterministic for simplicity. Also, whilst in this paper we will use, for simplicity, the same inductive bias for all protected characteristics, it is possible to instead use different inductive biases for different protected characteristics.

Given some policy $\inp\in\pols$ we define $\expr(\inp)$ to be the set of all $\iwfh\in[0,1]^{\prcs\times\exs}$ such that for all $(\prc,\arc)\in\ins$ and $\ac\in\acs$ we have:
\be
\inp(\prc,\arc,\ac)=\sum_{\ex\in\exs}\indi{\ex(\arc)=\ac}\iwfh(\prc,\ex)
\ee
noting that for all $\prc\in\prcs$ we have that $\iwfh(\prc,\blank)\in\ibs$. Note that $\expr(\inp)$ is always non-empty.

\subsection{Hedge and Bayesian Inference}\label{habisec}

Our meta-algorithm \alg\ takes, as input, an implementation of an instance of the classic \textsc{Hedge} algorithm. We shall call this \textsc{Hedge} implementation, which has an inherent inductive bias $\iwf\in\ibs$ and \emp{learning rate} $\lr>0$,  the \emp{base algorithm}. \textsc{Hedge} solves the following problem, which is a game between Nature and Learner proceeding in $\ntr$ trials. On each trial $t\in[\ntr]$ the following happens:
\begin{enumerate}
\item Nature selects a context $\cont{t}\in\cons$ and reveals it to Learner.
\item Learner selects an action $\act{t}\in\acs$ and reveals it to Nature.
\item Nature selects a \emp{loss} function $\lost{t}\in[0,1]^\acs$ and reveals it to Learner.
\end{enumerate}
The aim of Learner is to minimise the \emp{cumulative loss}:
\be
\sum_{t\in[\ntr]}\lost{t}(\act{t})
\ee
\textsc{Hedge} implicitly maintains a distribution $\iwfp\in\ibs$, initialised equal to $\iwf$. It has the following two subroutines:
\begin{itemize}
\item \qus\ takes, as input, a context $\arc\in\cons$ and returns a distribution $\acd\in\splx{\acs}$ defined such that for all $\ac\in\acs$ we have:
\be
\acd(\ac):=\sum_{\ex\in\exs}\indi{\ex(\arc)=\ac}\iwfp(\ex)
\ee
For some problems only certain contexts are possible on a given trial so, on such a trial, \qus\ only needs be callable for such contexts.
\item \ups\ takes, as input, a pair $(\arc,\arl)\in\cons\times\mathbb{R}^\acs$ then, for all $\ex\in\exs$, implicitly sets:
\be
\iwfp(\ex)\gets\iwfp(\ex)\exp(-\lr\arl(\ex(\arc)))
\ee
and finally implicitly normalises $\iwfp$.
\end{itemize}
\textsc{Hedge} solves the above problem by, on each trial $t\in[\ntr]$, doing the following. It first computes: 
\be
\acdt{t}\gets\qus(\cont{t})
\ee 
and samples $\act{t}$ from $\acdt{t}$. On receipt of $\lost{t}$ it then runs:
\be
\ups(\cont{t},\lost{t})
\ee
We note that any algorithm for discrete bayesian inference can be used as an algorithm for implementing \textsc{Hedge}. The contexts are equivalent to the random variables, which can each take a value in $\acs$, and $\iwf$ is the prior distribution. \textsc{Query} performs inference on a random variable and \textsc{Update} incorporates evidence, via Bayes' rule, about a specific random variable. Hence, if there exists no efficient exact implementation for our inductive bias we could use, instead, an efficient algorithm for approximate bayesian inference.

\section{Main Results}
We now describe the problems that we solve, as well as stating the performance of \alg.

\subsection{Fair Bandits}

We first introduce the \emp{fair bandit problem}, noting that any algorithm for this problem can also be used for fair full-information online learning, with the learning rate being automatically tuned via a \textsc{Hedge}-based doubling trick. We will show in Section \ref{redusec} that other problems can be efficiently reduced to this problem, such as those involving infinite context spaces, dynamic regret, and learning the distribution with which to have statistical parity with respect to. These reductions exploit the fact that, in what follows, the sequence $\seq{\edt{t}}{t\in[\ntr]}$ need not be constant.

The fair bandit problem is a game between Nature and Learner and runs in $\ntr$ trials. A-priori Nature selects a sequence:
\be
\seq{(\edt{t},\prct{t},\cont{t},\lost{t})}{t\in[\ntr]}
\ee
where $\edt{t}\in\cds$, $(\prct{t},\cont{t})\in\ins$ and $\lost{t}\in[0,1]^\acs$. However, Nature does not reveal the sequence to Learner. On each trial $t\in[\ntr]$ the following happens:
\begin{enumerate}
\item Nature reveals $\edt{t}$ to Learner.
\item Learner implicitly chooses some fair policy $\polt{t}\in\sps{\edt{t}}$.
\item Nature reveals $(\prct{t},\cont{t})$ to Learner.
\item Learner samples an action $\act{t}$ from $\polt{t}(\prct{t},\cont{t},\blank)$
\item Nature reveals $\lost{t}(\act{t})$ to Learner.
\end{enumerate}
Learner's goal is to minimise the \emp{cumulative loss}:
\be
\sum_{t\in[\ntr]}\lost{t}(\act{t})
\ee
To analyse our performance we define, for all $\comp\in\bigcap_{t\in[\ntr]}\sps{\edt{t}}$, our \emp{regret} as:
\be
\reg{\comp}:=\mathbb{E}\left[\sum_{t\in[\ntr]}\sum_{\ac\in\acs}(\polt{t}(\prct{t},\cont{t},\ac)-\comp(\prct{t},\cont{t},\ac))\arl(\ac)\right]
\ee
where the expectation is with respect to the draw of the actions $\seq{\act{t}}{t\in[\ntr]}$ which influence Learner's behaviour. We note that the regret $\reg{\comp}$ is the expected difference between our cumulative loss and that obtained by always choosing policy $\comp$.

To give our result we first define, for all $t\in[\ntr]$ and $\prc\in\prcs$, the \emp{support}:
\be
\supp{t}(\prc):=\{\arc\in\cons\,|\,\edt{t}(\prc,\arc)\neq0\}
\ee
For simplicity we will assume that $\cont{t}\in\supp{t}(\prct{t})$ although this is not actually necessary.

We have the following result:
\begin{theorem}\label{fbanth1}
Given a base algortihm with inductive bias $\iwf\in\ibs$ and learning rate $\lr:=\pra/\sqrt{\nma\ntr}$ for some $\pra>0$, \alg\ gives us the following regret for the fair bandit problem. For any fair policy: 
\be
\comp\in\bigcap_{t\in[\ntr]}\sps{\edt{t}}
\ee 
and any $\iwfh\in\expr(\comp)$ we have:
\be
\reg{\comp}\in\mathcal{O}\left(\left(\pra+\frac{\rwcs}{\pra}\right)\sqrt{\nma\ntr}\right)
\ee
where:
\be
\rwcs:=\sum_{\prc\in\prcs}\sum_{\ex\in\exs}\iwfh(\prc,\ex)\ln\left(\frac{\iwfh(\prc,\ex)}{\iwf(\ex)}\right)
\ee
The asymptotic time complexity of each trial $t\in[\ntr]$ is that of calling, for each $\prc\in\prcs$, $\qus(\arc)$ for each $\arc\in\supp{t}(\prc)$ and then calling $\ups(\arc,\blank)$ for each $\arc\in\supp{t}(\prc)$. Note that these multiple calls can typically be implemented in parallel much faster than the sequence of single calls. The space complexity is that of $\nmg$ instances of the base algorithm.
\end{theorem}

Note that, relative to any fair comparator policy, our regret is asymptotically identical to that of (the unfair strategy of) running \textsc{Exp4} with inductive bias $\iwf$ for each protected characteristic. We now give a couple of illustrative examples. Our first example is when the context set is unstructured (i.e. there is effectively no inductive bias). Such a context set could be generated by an a-priori clustering of an infinite context set. Here our base algorithm is that of running \textsc{Hedge} for each context independently. Note that, although the base algorithm treats each context independently, \alg\ does not. Direct from the bound of \textsc{Hedge} we see that the value of $\rwcs$ appearing in Theorem \ref{fbanth1} is:
\be
\rwcs\in\mathcal{O}(\nmg\nmc\ln(\nma))
\ee
Our second example is when the contexts are vertices of a graph. Here we follow some of the methodology in \cite{Herbster2021AGO}. We use the pre-processing step developed in \cite{Herbster2008OnlinePO, CesaBianchi2010RandomST} by sampling a uniformly random spanning tree and then linearising it via a depth first search. As our base algorithm we use the \emp{forward-backward algorithm} (for bayesian inference in Markov chains) over this linearisation. Our performance is then as follows. Suppose we have, for every $\prc\in\prcs$ a clustering of the graph such that $\comp(\prc,\arc,\blank)=\comp(\prc,\arc',\blank)$ if $\arc$ and $\arc'$ are in the same cluster. Let $\prwcs{\prc}$ be the \emp{effective resistance weighted cutsize} of this clustering. Then the expected value of $\rwcs$ appearing in Theorem \ref{fbanth1} is:
\be
\mathbb{E}[\rwcs]\in\mathcal{O}\left(\ln(\nmc\nma)\sum_{\prc\in\prcs}\prwcs{\prc}\right)
\ee
The per-trial time and space complexity of both these examples is only $\mathcal{O}(\nmg\nmc\nma)$.

\subsection{Fair Classification}
Once we have an algorithm for the fair bandit problem we will also be able to solve the following \emp{fair classification} problem. In this problem we have a known target $\apd\in\cds$ and an unknown probability distribution $\bcpd$ over the set: 
\be
\prcs\times\cons\times[0,1]^\acs
\ee
For some $\ntr\in\nat$ a sequence: 
\be
\seq{(\prct{t},\cont{t},\lost{t})}{t\in[\ntr]}
\ee
is drawn i.i.d. from $\bcpd$ and revealed to us. Our goal is to construct a fair policy $\inp\in\sps{\apd}$ that approximately minimises the \emp{generalisation error}:
\be
\mathbb{E}_{(\prc,\arc,\arl)\sim\bcpd}\left[\sum_{\ac\in\acs}\inp(\prc,\arc,\ac)\arl(\ac)\right]
\ee
To analyse our performance we define, for any $\comp\in\sps{\apd}$, the \emp{generalisation regret}:
\be
\genr(\comp):=\mathbb{E}_{(\prc,\arc,\arl)\sim\bcpd}\left[\sum_{\ac\in\acs}(\inp(\prc,\arc,\ac)-\comp(\prc,\arc,\ac))\arl(\ac)\right]
\ee
which is the difference between our generalisation error and that which would have been incurred if we had chosen $\comp$ instead of $\inp$.

We have the following result:
\begin{theorem}\label{fbclath}
Given that the base algorithm has inductive bias $\iwf\in\ibs$\,, we can utilise \alg\ to give us the following generalisation regret for the fair classification problem. For any fair policy $\comp\in\sps{\apd}$ and any $\iwfh\in\expr(\comp)$ we have:
\be
\mathbb{E}[\genr(\comp)]\in\tilde{\mathcal{O}}\left(\sqrt{\rwcs\nma/\ntr}\right)
\ee
where:
\be
\rwcs:=\sum_{\prc\in\prcs}\sum_{\ex\in\exs}\iwfh(\prc,\ex)\ln\left(\frac{\iwfh(\prc,\ex)}{\iwf(\ex)}\right)
\ee
The asymptotic per-trial time complexity is $\nmg\ntr$ times that of calling $\qus(\arc)$ for each $\arc\in\cons$ and then calling $\ups(\arc,\blank)$ for each $\arc\in\cons$. Note that these multiple calls can typically be implemented in parallel much faster than the sequence of single calls. The space complexity is that of $\nmg$ instances of the base algorithm.
\end{theorem}

We note that, relative to any fair comparator policy, the generalisation regret bound given above is just an $\mathcal{O}(\sqrt{\nma})$ factor off that of (the unfair strategy of) running \textsc{Hedge} for each protected characteristic and then applying online-to-batch conversion.

\section{The Meta-Algorithm}\label{algovvsec}

Here we given an overview of \alg\ and its analysis. We note that many statements in this section are far from immediate, the full analysis being deferred to Appendix \ref{alansec}. The pseudocode of \alg\ is given in Appendix \ref{pseudocsec}. At the end of this section we shall show how to apply \alg\ to the fair classification problem. 

Let $\pra$, $\lr$, $\comp$, $\iwf$ and $\iwfh$ be as in Theorem \ref{fbanth1}. Without loss of generality we can assume that:
\be
\pra\leq\sqrt{\ntr/\nma}
\ee
since the regret is never greater than $\ntr$. \alg\ maintains an instance of the base algorithm for each protected characteristic.
 For each $\prc\in\prcs$ let $\qsc{\prc}$ and $\usc{\prc}$ be the \qus\ and \ups\ subroutines for the instance of the base algorithm associated with the protected characteristic $\prc$. We now discuss the (implicit) operation of \alg\ on trial $t$.

\subsection{Definitions}

We first make some fundamental definitions. Firstly, given some $\prc\in\prcs$, recall that the instance of the base algorithm for protected characteristic $\prc$ implicitly maintains a distribution in $\ibs$. Let $\iwt{t}(\prc,\blank)$ be the value of this distribution at the start of trial $t$, noting that $\iwt{1}(\prc,\blank)=\iwf$. Define: 
\be
\eps:=\mathbb{R}^{\prcs\times\cons\times\acs}
\ee 
Given $(\cpv,\prc,\ac)\in\eps\times\prcs\times\acs$ define:
\be
\gamn{t}(\cpv,\prc,\ac):=\sum_{\arc\in\cons}\edt{t}(\prc,\arc)\cpv(\prc,\arc,\ac)
\ee
and define:
\be
\edn{t}(\cpv,\prc,\ac):=\max_{\prc'\in\prcs}\gamn{t}(\cpv,\prc',\ac)-\gamn{t}(\cpv,\prc,\ac)
\ee
Given $\cpv\in\eps$ define:
\be
\cer{t}(\cpv):=\sum_{\ac\in\acs}\max_{\prc\in\prcs}\edn{t}(\cpv,\prc,\ac)
\ee
Finally define, for all $(\cpv,\ac)\in\eps\times\acs$\,, the quantities:
\be
\kapu{t}(\cpv,\ac):=\operatorname{argmax}_{\prc\in\prcs}\gamn{t}(\cpv,\prc,\ac)
\ee
\be
\kapl{t}(\cpv,\ac):=\operatorname{argmin}_{\prc\in\prcs}\gamn{t}(\cpv,\prc,\ac)
\ee
where ties are broken arbitrarily.

\subsection{The Construction of the Policy}

We now describe how our fair policy $\polt{t}$ is constructed. We first construct an (unfair) policy $\rpol{t}\in\pols$ such that for all $(\prc,\arc)\in\ins$ we have: 
\be
\rpol{t}(\prc,\arc,\blank)\gets\qsc{\prc}(\arc)
\ee
although this only needs to be computed for $\arc\in\supp{t}(\prc)$.
Note that for all $(\prc,\arc,\ac)\in\prcs\times\cons\times\acs$ we have:
\begin{equation}\label{rpoleq1}
\rpol{t}(\prc,\arc,\ac)=\sum_{\ex\in\exs}\indi{\ex(\arc)=\ac}\iwt{t}(\prc,\ex)
\end{equation}
Now that we have $\rpol{t}$ we can define, for all $\ac\in\acs$ and $\prc\in\prcs$\,, the quantities:
\be
\cdin{t}(\prc,\ac)\la\edn{t}(\rpol{t},\prc,\ac)~~~~~;~~~~~\cra{t}\la\cer{t}(\rpol{t})
\ee 
We then define a function $\imp{t}\in\eps$ such that for all $(\prc,\arc,\ac)\in\prcs\times\cons\times\acs$ we have:
\be
\imp{t}(\prc,\arc,\ac)\la\frac{\rpol{t}(\prc,\arc,\ac)+\cdin{t}(\prc,\ac)}{1+\cra{t}}
\ee
and finally, for all $(\prc,\arc,\ac)\in\prcs\times\cons\times\acs$ we define:
\be
\polt{t}(\prc,\arc,\ac)\la\imp{t}(\prc,\arc,\ac)+\frac{1}{\nma}\left(1-\sum_{\ac'\in\acs}\imp{t}(\prc,\arc,\ac')\right)
\ee
We then have that $\polt{t}\in\sps{\spt{t}}$ as required. In addition to this, $\polt{t}$ has the following crucial properties. For all $(\prc,\arc,\ac)\in\prcs\times\cons\times\acs$ we have:
\begin{equation}\label{imeq4baeq}
\polt{t}(\prc,\arc,\ac)\geq\frac{\rpol{t}(\prc,\arc,\ac)}{1+\cra{t}}
\end{equation}
and for all $(\prc,\arc)\in\ins$ we have:
\begin{equation}\label{difbobybeeq}
\sum_{\ac\in\acs}\max\{0,\polt{t}(\prc,\arc,\ac)-\rpol{t}(\prc,\arc,\ac)\}\leq\cra{t}
\end{equation}

\subsection{The Update}
We now describe how our instances of the base algorithm are updated at the end of trial $t$. We will first define a convex objective function $\slf{t}:\eps\rightarrow\mathbb{R}$ which will never actually be known by Learner. Our objective function is defined such that for all $\cpv\in\eps$ we have:
\be
\slf{t}(\cpv):=\indi{\cra{t}\leq1}\sum_{\ac\in\acs}\loft{t}(\ac)\cpv(\prct{t},\cont{t},\ac)+\cer{t}(\cpv)
\ee
The reason we have chosen such an objective function is due to the following two properties, where the derivation of Equation \eqref{sulflem1eq} uses Equation \eqref{difbobybeeq}.
\begin{equation}\label{sulflem1eq}
\slf{t}(\rpol{t})\geq\sum_{\ac\in\acs}\polt{t}(\prct{t},\cont{t},\ac)\loft{t}(\ac)
\end{equation}
\begin{equation}\label{sulflem2eq}
\slf{t}(\comp)\leq\sum_{\ac\in\acs}\comp(\prct{t},\cont{t},\ac)\loft{t}(\ac)
\end{equation}
A sub-gradient of $\slf{t}$ at any point $\cpv\in\eps$ is given by $\tgr{t}(\cpv,\blank,\blank,\blank)$ where we define $\tgr{t}:\eps\times\prcs\times\cons\times\acs\rightarrow\real$ such that for all $(\cpv,\prc,\arc,\ac)\in\eps\times\prcs\times\cons\times\acs$ we have:
\begin{align*}
\tgr{t}(\cpv,\prc,\arc,\ac):=&\indi{\cra{t}\leq1}\indi{(\prc,\arc)=(\prct{t},\cont{t})}\loft{t}(\ac)\\
&+\indi{\prc=\kapu{t}(\cpv,\ac)}\edt{t}(\prc,\arc)\\
&-\indi{\prc=\kapl{t}(\cpv,\ac)}\edt{t}(\prc,\arc)
\end{align*}
Due to the fact that we don't know the entire function $\lost{t}$\,, we can't compute the sub-gradient at $\rpol{t}$ so instead we will borrow the technique, from \textsc{Exp4}, of using $\indi{\ac=\act{t}}\loft{t}(\act{t})/\polt{t}(\prct{t},\cont{t},\act{t})$ as an unbiased estimator of $\loft{t}(\ac)$. This gives us the following function $\gra{t}$ that we call the \emp{pseudo-gradient}. $\gra{t}\in\eps$ is defined such that for all $(\prc,\arc,\ac)\in\prcs\times\cons\times\acs$ we have:
\begin{align*}
\gra{t}(\prc,\arc,\ac):=&\indi{\cra{t}\leq1}\indi{(\prc,\arc,\ac)=(\prct{t},\cont{t},\act{t})}\frac{\loft{t}(\act{t})}{\polt{t}(\prct{t},\cont{t},\act{t})}\\
&+\indi{\prc=\kanu{t}(\ac)}\edt{t}(\prc,\arc)-\indi{\prc=\kanl{t}(\ac)}\edt{t}(\prc,\arc)
\end{align*}
where, for all $\ac\in\acs$\,, we have defined:
\be
\kanu{t}(\ac):=\kapu{t}(\rpol{t},\ac)~~~~~;~~~~~\kanl{t}(\ac):=\kapl{t}(\rpol{t},\ac)
\ee
$\gra{t}$ is an unbiased estimator of the sub-gradient $\tgr{t}(\rpol{t},\blank,\blank,\blank)$, in that it is equal to it in expectation over the draw of $\act{t}$. Note that for the full-information online and batch classification versions of \alg\ we can use the true sub-gradient instead of $\gra{t}$ in order to give a deterministic algorithm. 

We now give some important properties of $\gra{t}$. First define the \emp{instantaneous regret} as:
\be
\inr{t}:=\sum_{\ac\in\acs}(\polt{t}(\prct{t},\cont{t},\ac)-\comp(\prct{t},\cont{t},\ac))\loft{t}(\ac)
\ee
Since $\gra{t}$ is an unbiased estimator of a sub-gradient of $\slf{t}$ at $\rpol{t}$, we have, by equations \eqref{sulflem1eq} and \eqref{sulflem2eq}, that:
\begin{align}
\notag&\mathbb{E}\left[\sum_{\prc\in\prcs}\sum_{\arc\in\cons}\sum_{\ac\in\acs}(\rpol{t}(\prc,\arc,\ac)-\comp(\prc,\arc,\ac))\gra{t}(\prc,\arc,\ac)\,\Bigg|\,\rpol{t}\right]\\
\label{hdganlem1eq}&\geq\inr{t}
\end{align}
Now define $\geft{t}:\prcs\times\exs\rightarrow\mathbb{R}$ such that for all $(\prc,\ex)\in\prcs\times\exs$ we have:
\be
\geft{t}(\prc,\ex):=\sum_{\arc\in\cons}\gra{t}(\prc,\arc,\ex(\arc))
\ee
Equations \eqref{rpoleq1} and \eqref{hdganlem1eq}\,, and the fact that $\iwfh\in\expr(\comp)$\,, combine to give us:
\begin{equation}\label{hdganlem1.5eq}
\mathbb{E}\left[\sum_{\prc\in\prcs}\sum_{\ex\in\exs}(\iwt{t}(\prc,\ex)-\iwfh(\prc,\ex))\geft{t}(\prc,\ex)\Bigg|\,\iwt{t}\right]\geq\inr{t}
\end{equation}

The reason that the term $\indi{\cra{t}\leq1}$ appears in the definition of our objective function $\slf{t}$ is to ensure the following crucial property, which follows from equations \eqref{rpoleq1} and \eqref{imeq4baeq}.

\begin{equation}\label{hdganlem2eq}
\mathbb{E}\left[\sum_{\prc\in\prcs}\sum_{\ex\in\exs}\iwt{t}(\prc,\ex)\geft{t}(\prc,\ex)^2\,\Bigg|\,\iwt{t}\right]\leq 8\nma
\end{equation}
In addition, for all $(\prc,\ex)\in\prcs\times\exs$ we have:
\begin{equation}\label{bbbmkeq}
\geft{t}(\prc,\ex)\geq-\nma
\end{equation}
To update we run, for each $\prc\in\prcs$ and $\arc\in\supp{t}(\prc)$\,, the function: 
\be
\usc{\prc}(\arc,\gra{t}(\prc,\arc,\blank))
\ee 
This update ensures that for all $(\prc,\ex)\in\prcs\times\exs$ we have:
\begin{equation}\label{rohulem1eq}
\iwt{t+1}(\prc,\ex)=\frac{\iwt{t}(\prc,\ex)\exp(-\lr\geft{t}(\prc,\ex))}{\sum_{\ex'\in\exs}\iwt{t}(\prc,\ex')\exp(-\lr\geft{t}(\prc,\ex'))}
\end{equation}
We define the \emp{relative entropy} $\re:\ibs\times\ibs\rightarrow\mathbb{R}$ such that for all $\avt,\avt'\in\ibs$ we have:
\be
\re(\avt,\avt'):=\sum_{\ex\in\exs}\avt(\ex)\ln\left(\frac{\avt(\ex)}{\avt'(\ex)}\right)
\ee
Using Equation \eqref{bbbmkeq}, we apply the first part of the \textsc{Hedge} analysis (but modified to take into account that $\gra{t}$ can have negative values) to Eqaution \eqref{rohulem1eq}. For all $\prc\in\prcs$ this gives us:
\begin{align*}
&\re(\iwfh,\iwt{t}(\prc,\blank))-\re(\iwfh,\iwt{t+1}(\prc,\blank))\\
\geq~&\lr\sum_{\ex\in\exs}(\iwt{t}(\prc,\ex)-\iwfh(\prc,\ex))\geft{t}(\prc,\ex)\\
&-\lr^2\sum_{\ex\in\exs}\iwt{t}(\prc,\ex)\geft{t}(\prc,\ex)^2
\end{align*}
This equation combines with equations \eqref{hdganlem1.5eq} and \eqref{hdganlem2eq} to give us:
\begin{align}
\notag&\sum_{\prc\in\prcs}\mathbb{E}\left[\re(\iwfh,\iwt{t}(\prc,\blank))-\re(\iwfh,\iwt{t+1}(\prc,\blank))\,|\,\iwt{t}\right]\\
\label{flftlem1eq}&\geq\lr\inr{t}-8\lr^2\nma
\end{align}
This completes the description and analysis overview of trial $t$. 

\subsection{Results}

To get the overall regret bound we take expectations on Equation \eqref{flftlem1eq} and sum over all $t\in[\ntr]$\,, which gives us:
\begin{equation}\label{fb4blemeq}
\reg{\comp}\leq\left(8\pra+\frac{\rwcs}{\pra}\right)\sqrt{\nma\ntr}
\end{equation}
where $\rwcs$ is as in Theorem \ref{fbanth1}.
Note that in the full-information problem we can maintain $\mathcal{O}(\ln(\ntr))$ copies of \alg\ with exponentially increasing learning rates, and combine them together with \textsc{Hedge}. We can also do this for any parameters in the inductive bias. Hence, with full-information we have no free parameters. To obtain the fair classification bounds we simply run this full-information version of \alg\ online over the training set with $\edt{t}=\apd$ for all $t\in[\ntr]$, and then output:
\be
\inp:=\frac{1}{\ntr}\sum_{t\in[\ntr]}\polt{t}
\ee
which can be computed online. Note that this is effectively the online-to-batch conversion process given in \cite{CesaBianchi2001OnTG} so the bound in Theorem \ref{fbclath} comes directly from that paper and Equation \eqref{fb4blemeq}.

\nc{\tcons}{\mathcal{X}^*}
\nc{\trpd}[1]{\mathcal{\mu}^*_{#1}}
\nc{\trpdc}{\mathcal{\mu}^*}
\nc{\tcds}{\cds^*}
\nc{\tpols}{\pols^*}
\nc{\tsps}{\mathcal{F}}
\nc{\tpolt}[1]{\pi^*_{#1}}
\nc{\tcp}[1]{\tilde{\pi}^*_{#1}}
\nc{\tarc}{\arc^*}
\nc{\xst}[1]{x^*_{#1}}
\nc{\ascb}{\mathcal{U}}
\nc{\arcs}{x^*}
\nc{\rot}{r}
\nc{\psbt}[1]{\mathcal{X}'(#1)}
\nc{\cpols}{\tilde{\pi}^*}
\nc{\pos}{p}
\nc{\msap}{\tilde{\mu}^*}

\section{Reductions}\label{redusec}

We now give some important problems that are not immediately instances of the fair bandit problem as we described it, but reduce to it instead. In all these problems we denote the true context space by $\tcons$ and reduce it to a context set $\cons$ for our algorithm to be applied. Let $\xst{t}$ be the (true) context in $\tcons$ that is revealed to us on trial $t$. We define the sets $\tcds$ and $\tpols$ to be equivalent to the sets $\cds$ and $\pols$ respectively but using $\tcons$ instead of $\cons$. We define $\tsps:\tcds\rightarrow2^{\tpols}$ as we did for the set $\cons$.

\subsection{Empirical Fairness and Dynamic Environments}\label{eradessec}
We first study the problem of always maintaining statistical parity with respect to the empirical distribution of the instances seen so far (or any sequence of targets), as well as the problem of learning in a dynamic environment. The general problem considered here is as follows. The context set $\tcons$ is unstructured in that there is effectively no inductive bias on it, noting that such a context set could be generated by an a-priori clustering of a structured context space. At the start of each trial $t\in[\ntr]$ we are given some $\trpd{t}\in\tcds$ and must draw our action from a policy $\tpolt{t}\in\tsps(\trpd{t})$. We wish to bound our performance with respect to any sequence $\seq{\tcp{t}}{t\in[\ntr]}$ in which $\tcp{t}\in\tsps(\trpd{t})$ for all $t\in[\ntr]$. In order to handle this non-stationarity we utilise, as our base algorithm, \textsc{FixedShare} \cite{Herbster1995TrackingTB} for each context independently. Note that, although the base algorithm treats each context independently, \alg\ does not. \textsc{FixedShare} works via defining $\cons:=\tcons\times[\ntr]$ and choosing $\cont{t}:=(\xst{t},t)$ for all $t\in[\ntr]$. For all $t\in[\ntr]$ we define $\edt{t}$ such that for all $(\prc,\tarc, t')\in\prcs\times\tcons\times[\ntr]$ we have:
\be
\edt{t}(\prc, (\tarc, t')):=\indi{t'=t}\trpd{t}(\prc,\tarc)
\ee 
With \textsc{FixedShare} as our base algorithm, \alg\ runs in a time of only $\mathcal{O}(\nmg|\tcons|\nma)$ per trial. Directly from \cite{Herbster2001TrackingTB} we have that the value of $\rwcs$ appearing in Theorem \ref{fbanth1} is:
\be
\nmg\nmc+\sum_{t\in[\ntr-1]}\sum_{\prc\in\prcs}\sum_{\arcs\in\tcons}\sum_{\ac\in\acs}|\tcp{t+1}(\prc,\arcs,\ac)-\tcp{t}(\prc,\arcs,\ac)|
\ee
up to factors logarithmic in $\nma$ and $\ntr$. In some situations we will be required to be \emp{empirically fair}, in that $\trpd{t}$ is defined by the empirical distribution of the instances seen so far (where we exclude protected characteristics as yet unseen).

\subsection{Massive and Infinite Context Spaces}
We now describe how to utilise the methodology of \cite{Pasteris2023NearestNW} to efficiently handle massive and infinite context spaces which can be hierarchically decomposed (such as euclidean space). Here the set $\cons$ is the vertex set of a full binary tree where each element of $\cons$ is a subset of $\tcons$. The root of $\cons$, which is denoted by $\rot$, is equal to the entire set $\tcons$. For each internal vertex $\arc\in\cons$ let $\lch{\arc}$ and $\rch{\arc}$ be its two children. We have that $\lch{\arc}\cap\rch{\arc}=\emptyset$ and $\lch{\arc}\cup\rch{\arc}=\arc$. It is the action of a hierarchical decomposition algorithm to partition $\arc$ into $\lch{\arc}$ and $\rch{\arc}$. We assume that we have a known target $\trpdc\in\tcds$ with which to maintain statistical parity with respect to, and that we can compute: 
\be
\trpdc(\prc,\arc):=\int_{\arcs\in\arc}\trpdc(\prc,\arcs)
\ee 
for all $\prc\in\prcs$ and $\arc\in\cons$. Our base algorithm will not explicitly maintain $\cons$, which can be massive. Instead, our base algorithm maintains, for each $\prc\in\prcs$, a dynamic (i.e. changing from trial to trial) subtree $\psbt{\prc}$ of $\cons$ which contains $\rot$. These subtrees are initialised so that they contain $\rot$ as a single vertex. On each trial $t\in[\ntr]$ we define $\edt{t}$ as follows. For each $\prc\in\prcs$ and $\arc\in\cons$ we have $\edt{t}(\prc,\arc)=\trpdc(\prc,\arc)$ if $\arc$ is a leaf of $\psbt{\prc}$ and $\edt{t}(\prc,\arc)=0$ otherwise. We choose $\cont{t}$ to be the unique leaf of $\psbt{\prct{t}}$ that contains $\xst{t}$.  At the end of trial $t$ we grow the subtree $\psbt{\prct{t}}$ by adding to it the children of $\cont{t}$ if they exist. Our base algorithm follows the simplest algorithm of \cite{Pasteris2023NearestNW}, which only requires us to perform, for each $\prc\in\prcs$, belief propagation \cite{Pearl1982ReverendBO} over the subtree $\psbt{\prc}$ rather than over the entire tree $\cons$. The running time of \alg\ is then only $\mathcal{O}(\nma\ntr)$ per trial. Our performance guarantees follow from Theorem \ref{fbanth1} and the performance guarantees in \cite{Pasteris2023NearestNW}.

\subsection{Approximate Statistical Parity with IID Contexts}\label{aspwicssec}
Finally we consider the problem of maintaining approximate statistical parity on all trials (with high probability) with respect to an unknown distribution, when the instances are drawn from it i.i.d. We assume that the context space can be hierarchically decomposed as in the previous subsection. Formally, we have some $\trpd{}\in\tcds$ such that on each trial $t\in[\ntr]$ the context $\xst{t}$ is drawn from $\trpd{}(\prct{t},\blank)$. Our aim is to, with high probability, approximate statistical parity with respect to $\trpd{}$ on all trials $t\in[\ntr]$. To do this we follow the algorithm of the previous subsection with a couple of modifications. We have some parameter $\nsm\in\nat$ which determines the quality of our approximation. The first modification is that, on any trial $t$, we only add the children of $\cont{t}$ to the subtree $\psbt{\prct{t}}$ if we have had at least $\nsm$ trials $s$ in which $\prct{s}=\prct{t}$ and $\xst{s}\in\cont{t}$. When then define $\pos(\prct{t},\lch{\cont{t}})$ and $\pos(\prct{t},\rch{\cont{t}})$ to be the proportion of these trials $s$ with $\xst{s}\in\lch{\cont{t}}$ and $\xst{s}\in\rch{\cont{t}}$ respectively. Finally, for any $\arc\in\cons$, we don't know $\trpdc(\prc,\arc)$ so approximate it with a value $\msap(\prc,\arc)$ which is recursively defined such that $\msap(\prc,\rot)=1$ and when $\arc\in\cons\setminus\{\rot\}$ we have $\msap(\prc,\arc)=\pos(\prc,\arc)\msap(\prc,\pnt{\arc})$ where $\pnt{\arc}$ is the parent of $\arc$. In Appendix \ref{afltdwicsec} we analyse the quality of the approximation of statistical parity. Again, our performance guarantees follow from Theorem \ref{fbanth1} and the performance guarantees in \cite{Pasteris2023NearestNW}.

\section*{Acknowledgements}
Research funded by the Defence Science and Technology Laboratory (Dstl) which is an executive agency of the UK Ministry of Defence providing world class expertise and delivering cutting-edge science and technology for the benefit of the nation and allies. The research supports the Autonomous Resilient Cyber Defence (ARCD) project within the Dstl Cyber Defence Enhancement programme.


\bibliography{Fairbib}
\bibliographystyle{icml2025}

\newpage
\appendix
\onecolumn

\section{Additional Related Work}\label{arwsec}

There has been much work on various types of fairness with respect to actions, rather than instances, in i.i.d. stochastic bandit problems. For example, \cite{Joseph2016FairnessIL} considered the problem of ensuring that a better action (one with lower mean loss) never has a lower probability of being chosen than a worse action (one with higher mean loss), \cite{Wang2021FairnessOE} considered the problem of choosing each action with a probability proportional to its \emp{merit}, \cite{Liu2017CalibratedFI} considered the problem of treating similar actions similarly, and \cite{Patil2019AchievingFI} considered the constraint of selecting each action at least a pre-specified number of times. A form of fairness with respect to instances was studied in \cite{Huang2020AchievingUF} which considered the constraint of equalising the cumulative mean loss of all protected groups (all instances that have some given protected characteristic) in stochastic linear bandits. We note that, whilst this constraint will be desired in some situations, it is very different from statistical parity. Possibly the closest work to ours is \cite{Bechavod2019EqualOI}, which considered the i.i.d. stochastic online binary classification problem (classifying contexts as positive or negative) under the constraint that false negatives are equalised over two protected groups, under the assumption of the existence of an efficient oracle for empirical risk minimisation over the hypothesis space. They specifically considered a partial feedback setting where the true class is only revealed when the positive class is chosen. However, the fairness constraints are only approximated and the algorithm is relatively inefficient: taking a per-trial time that is polynomial in the number of trials (the exponent is not given in the paper). Also, empirical risk minimisation does not lead to the bounds of \textsc{Exp4}.

\section{Pseudocode}\label{pseudocsec}
Here we give the pseudocode for \alg\ on trial $t$. Note that in the pseudocode, for all $\ac\in\acs$ we have $\psih{t}(\ac):=\imp{t}(\prct{t},\cont{t},\ac)$, and $\pih{t}(\ac)=\polt{t}(\prct{t},\cont{t},\ac)$ where $\imp{t}$ is as defined in Section \ref{algovvsec}. Note also that for all $\prc\in\prcs$ we maintain an instance of the base algorithm with subroutines denoted by $\qsc{\prc}$ and $\usc{\prc}$. On trial $t$ \alg\ does the following.
\newline
\begin{algorithmic}
\State Receive $\edt{t}, \cont{t}$ and $\prct{t}$
\For{$\prc\in\prcs$}
\For{$\arc\in\supp{t}(\prc)$}
\State $\rpol{t}(\prc,\arc,\blank)\gets\qsc{\prc}(\arc)$
\For{$\ac\in\acs$}
\State $\omwt{t}(\prc,\ac)\gets\sum_{\arc\in\supp{t}(\prc)}\edt{t}(\prc,\arc)\rpol{t}(\prc,\arc,\ac)$
\EndFor
\EndFor
\EndFor
\For{$(\prc,\ac)\in\prcs\times\acs$}
\State $\cdin{t}(\prc,\ac)\gets\max_{\prc'\in\prcs}\omwt{t}(\prc',\ac)-\omwt{t}(\prc,\ac)$
\EndFor
\State $\cra{t}\gets\sum_{\ac\in\acs}\max_{\prc\in\prcs}\cdin{t}(\prc,\ac)$
\For{$\ac\in\acs$}
\State $\psih{t}(\ac)\gets(\rpol{t}(\prct{t},\cont{t},\ac)+\cdin{t}(\prct{t},\ac))/(1+\cra{t})$
\EndFor
\State $\hsu{t}\gets1-\sum_{\ac\in\acs}\psih{t}(\ac)$
\For{$\ac\in\acs$}
\State $\pih{t}(\ac)\gets\psih{t}(\ac)+\hsu{t}/\nma$
\EndFor
\State Draw $\act{t}$ from probability distribution $\pih{t}$
\State Receive $\loft{t}(\act{t})$
\For{$\ac\in\acs$}
\State $\kanu{t}(\ac):=\operatorname{argmax}_{\prc\in\prcs}\omwt{t}(\prc,\ac)$
\State $\kanl{t}(\ac):=\operatorname{argmin}_{\prc\in\prcs}\omwt{t}(\prc,\ac)$
\EndFor
\For{$\prc\in\prcs$}
\For{$\arc\in\supp{t}(\prc)$}
\For{$\ac\in\acs$}
\State
$\gra{t}(\prc,\arc,\ac)\gets\indi{\cra{t}\leq1}\indi{(\prc,\arc,\ac)=(\prct{t},\cont{t},\act{t})}\loft{t}(\act{t})/\pih{t}(\act{t})+\indi{\prc=\kanu{t}(\ac)}\edt{t}(\prc,\arc)-\indi{\prc=\kanl{t}(\ac)}\edt{t}(\prc,\arc)$
\EndFor
\State $\usc{\prc}(\arc,\gra{t}(\prc,\arc,\blank))$
\EndFor
\EndFor
\end{algorithmic}

\section{Analysis}\label{alansec}

Here we describe and analyse the mechanics of \alg. All lemmas stated in this section are proved in Section \ref{profsec}. 

Let $\pra$, $\comp$, $\iwf$ and $\iwfh$ be as in Theorem \ref{fbanth1}. Without loss of generality we can assume that:
\be
\pra\leq\sqrt{\ntr/\nma}
\ee
otherwise our regret bound would be vacuous. Recall that \alg\ takes as input an algorithm (a.k.a. the \emp{base algorithm}) for implementing \text{Hedge} with inductive bias $\iwf$. For the base algorithm we will choose the learning rate:
\be
\lr:=\frac{\pra}{\sqrt{\nma\ntr}}
\ee 
We will maintain $\nmg$ instances of this algorithm, one for each protected characteristic $\prc\in\prcs$. For each $\prc\in\prcs$ let $\qsc{\prc}$ and $\usc{\prc}$ be the \qus\ and \ups\ subroutines for the instance of the base algorithm associated with the protected characteristic $\prc$. Given some $\prc\in\prcs$, recall (from Section \ref{habisec}) that the instance of the base algorithm for protected characteristic $\prc$ implicitly maintains a distribution in $\ibs$. Let $\iwt{t}(\prc,\blank)$ be the value of this distribution at the start of trial $t$, so that $\iwt{1}(\prc,\blank)=\iwf$.

We now describe and analyse how \alg\ implicitly behaves on trial $t$ and, in doing so, will bound the \emp{instantaneous regret}:
\be
\inr{t}:=\sum_{\ac\in\acs}(\polt{t}(\prct{t},\cont{t},\ac)-\comp(\prct{t},\cont{t},\ac))\loft{t}(\ac)
\ee

\subsection{Fundamental Definitions}
We now make some fundamental definitions. We first define: 
\be
\eps:=\mathbb{R}^{\prcs\times\cons\times\acs}
\ee 
Given $(\cpv,\prc,\ac)\in\eps\times\prcs\times\acs$ define:
\be
\gamn{t}(\cpv,\prc,\ac):=\sum_{\arc\in\supp{t}(\prc)}\edt{t}(\prc,\arc)\cpv(\prc,\arc,\ac)=\sum_{\arc\in\cons}\edt{t}(\prc,\arc)\cpv(\prc,\arc,\ac)
\ee
and define:
\be
\edn{t}(\cpv,\prc,\ac):=\max_{\prc'\in\prcs}\gamn{t}(\cpv,\prc',\ac)-\gamn{t}(\cpv,\prc,\ac)
\ee
Given $\cpv\in\eps$ define:
\be
\cer{t}(\cpv):=\sum_{\ac\in\acs}\max_{\prc\in\prcs}\edn{t}(\cpv,\prc,\ac)
\ee
Finally, we define $\epl{t}$ to be the set of all $\cpv\in\eps$ in which for all $\prc,\prc'\in\prcs$ and $\ac\in\acs$ we have:
\be
\gamn{t}(\cpv,\prc,\ac)=\gamn{t}(\cpv,\prc',\ac)
\ee
We note that $\sps{\spt{t}}=\pols\cap\epl{t}$.

\subsection{Computing the Policy}\label{ppsec}
At the start of trial $t$ \alg\ constructs a policy $\rpol{t}\in\pols$ called the \emp{raw policy} such that for all $(\prc,\arc)\in\ins$ we have:
\be 
\rpol{t}(\prc,\arc,\blank)\gets\qsc{\prc}(\arc)
\ee
noting that we need only compute $\rpol{t}(\prc,\arc,\blank)$ if $\arc\in\supp{t}(\prc)$. We immediately have the following lemma:
\begin{lemma}\label{roqslem1}
For all $(\prc,\arc,\ac)\in\prcs\times\cons\times\acs$ we have:
\be
\rpol{t}(\prc,\arc,\ac)=\sum_{\ex\in\exs}\indi{\ex(\arc)=\ac}\iwt{t}(\prc,\ex)
\ee
\end{lemma}
We now describe the conversion of our raw policy $\rpol{t}$ into our fair policy $\polt{t}$. We call this conversion process \emp{policy processing}. First define, for all $\ac\in\acs$ and $\prc\in\prcs$\,, the quantities:
\be
\cdin{t}(\prc,\ac)\la\edn{t}(\rpol{t},\prc,\ac)~~~~~;~~~~~\cra{t}\la\cer{t}(\rpol{t})
\ee 
We define a function $\imp{t}\in\eps$ such that for any $(\prc,\arc,\ac)\in\prcs\times\cons\times\acs$ we have:
\be
\imp{t}(\prc,\arc,\ac)\la\frac{\rpol{t}(\prc,\arc,\ac)+\cdin{t}(\prc,\ac)}{1+\cra{t}}
\ee
The function $\imp{t}$ has the following properties:

\begin{lemma}\label{pieflem0}
We have that $\imp{t}\in\epl{t}$. In addition we have, for all $(\prc,\arc)\in\ins$, that:
\be
\sum_{\ac\in\acs}\imp{t}(\prc,\arc,\ac)\leq1
\ee
and that $\imp{t}(\prc,\arc,\ac)\geq0$ for all $\ac\in\acs$.
\end{lemma}

The above lemma shows that the function $\imp{t}$ satisfies most of our constraints. However, it is not necessarily in $\pols$ as the functions $\imp{t}(\prc,\arc,\blank)$ may not lie on the simplex $\splx{\acs}$. In order to convert it into $\polt{t}$ we simply add, to each of the functions $\imp{t}(\prc,\arc,\blank)$, some the uniform distribution. Specifically, for all $(\prc,\arc,\ac)\in\prcs\times\cons\times\acs$ we define:
\be
\polt{t}(\prc,\arc,\ac)\la\imp{t}(\prc,\arc,\ac)+\frac{1}{\nma}\left(1-\sum_{\ac'\in\acs}\imp{t}(\prc,\arc,\ac')\right)
\ee
The following lemma, which follows from Lemma \ref{pieflem0}, confirms that $\polt{t}$ does indeed lie in $\sps{\spt{t}}$.
\begin{lemma}\label{pieflem1}
We have:
\be
\polt{t}\in\sps{\spt{t}}
\ee
\end{lemma}

We now give two additional lemmas that are crucial to our analysis. The first lemma follows from Lemma \ref{pieflem0}.

\begin{lemma}\label{nwowrlem1}
For all $(\prc,\arc,\ac)\in\prcs\times\cons\times\acs$ we have:
\be
\polt{t}(\prc,\arc,\ac)\geq\frac{\rpol{t}(\prc,\arc,\ac)}{1+\cra{t}}
\ee
\end{lemma}

The next lemma follows from lemmas \ref{pieflem1} and \ref{nwowrlem1}

\begin{lemma}\label{disclem1}
For all $(\prc,\arc)\in\ins$ we have:
\be
\sum_{\ac\in\acs}\max\{0,\polt{t}(\prc,\arc,\ac)-\rpol{t}(\prc,\arc,\ac)\}\leq\cra{t}
\ee
\end{lemma}

We have now derived all the properties of $\polt{t}$ needed to proceed with the algorithm and analysis.

\subsection{The Objective Function and its Pseudo-Gradient}
Now that we have shown how to compute $\polt{t}$ we turn to the update of the base algorithm instances at the end of the trial. In order to do this we need a convex objective function $\slf{t}:\eps\rightarrow\mathbb{R}$ which we note will never actually be known by Learner. Our objective function is defined such that for all $\cpv\in\eps$ we have:
\be
\slf{t}(\cpv):=\indi{\cra{t}\leq1}\sum_{\ac\in\acs}\loft{t}(\ac)\cpv(\prct{t},\cont{t},\ac)+\cer{t}(\cpv)
\ee
The reason we have chosen such an objective function is due to the following two lemmas, where Lemma \ref{sulflem1} follows from Lemma \ref{disclem1}. We note that the appearance of $\indi{\cra{t}\leq1}$ in the definition of $\slf{t}$ is to ensure another crucial lemma, which we shall present later.

\begin{lemma}\label{sulflem1}
We have:
\be
\slf{t}(\rpol{t})\geq\sum_{\ac\in\acs}\polt{t}(\prct{t},\cont{t},\ac)\loft{t}(\ac)
\ee
\end{lemma}

\begin{lemma}\label{sulflem2}
We have:
\be
\slf{t}(\comp)\leq\sum_{\ac\in\acs}\comp(\prct{t},\cont{t},\ac)\loft{t}(\ac)
\ee
\end{lemma}

We will show that $\slf{t}$ is indeed convex by showing that it has a sub-gradient at every point in $\eps$. In order to give such a sub-gradient we first define, for all $(\cpv,\ac)\in\eps\times\acs$, the quantities:
\be
\kapu{t}(\cpv,\ac):=\operatorname{argmax}_{\prc\in\prcs}\gamn{t}(\cpv,\prc,\ac)~~~~~;~~~~~\kapl{t}(\cpv,\ac):=\operatorname{argmin}_{\prc\in\prcs}\gamn{t}(\cpv,\prc,\ac)
\ee
where ties are broken arbitrarily.
We then define $\tgr{t}:\eps\times\prcs\times\cons\times\acs\rightarrow\real$ such that for all $(\cpv,\prc,\arc,\ac)\in\eps\times\prcs\times\cons\times\acs$ we have:
\be
\tgr{t}(\cpv,\prc,\arc,\ac):=\indi{\cra{t}\leq1}\indi{(\prc,\arc)=(\prct{t},\cont{t})}\loft{t}(\ac)+\indi{\prc=\kapu{t}(\cpv,\ac)}\edt{t}(\prc,\arc)-\indi{\prc=\kapl{t}(\cpv,\ac)}\edt{t}(\prc,\arc)
\ee
The next lemma shows that for all $\cpv\in\eps$ we have that $\tgr{t}(\cpv,\blank,\blank,\blank)$ is a sub-gradient of $\slf{t}$ at $\cpv$.

\begin{lemma}\label{sulflem3}
For any $\cpv,\cpv'\in\eps$ we have:
\be
\slf{t}(\cpv)-\slf{t}(\cpv')\leq\sum_{\prc\in\prcs}\sum_{\arc\in\cons}\sum_{\ac\in\acs}\tgr{t}(\cpv,\prc,\arc,\ac)(\cpv(\prc,\arc,\ac)-\cpv'(\prc,\arc,\ac))
\ee
\end{lemma}

Due to the fact that we don't know the entire function $\lost{t}$\,, we can't compute the sub-gradient $\tgr{t}(\rpol{t},\blank,\blank,\blank)$ so instead, will borrow the technique, from \textsc{Exp4}, of using $\indi{\ac=\act{t}}\loft{t}(\act{t})/\polt{t}(\prct{t},\cont{t},\act{t})$ as an unbiased estimator of $\loft{t}(\ac)$. This gives us the following function $\gra{t}$ that we call the \emp{pseudo-gradient}. $\gra{t}\in\eps$ is defined such that for all $(\prc,\arc,\ac)\in\prcs\times\cons\times\acs$ we have:
\be
\gra{t}(\prc,\arc,\ac):=\indi{\cra{t}\leq1}\indi{(\prc,\arc,\ac)=(\prct{t},\cont{t},\act{t})}\frac{\loft{t}(\act{t})}{\polt{t}(\prct{t},\cont{t},\act{t})}+\indi{\prc=\kapu{t}(\rpol{t},\ac)}\edt{t}(\prc,\arc)-\indi{\prc=\kapl{t}(\rpol{t},\ac)}\edt{t}(\prc,\arc)
\ee
The following lemma states that $\gra{t}$ is an unbiased estimator of the sub-gradient $\tgr{t}(\rpol{t},\blank,\blank,\blank)$.

\begin{lemma}\label{sulflem4}
For all $(\prc,\arc,\ac)\in\prcs\times\cons\times\acs$ we have:
\be
\expt{\gra{t}(\prc,\arc,\ac)\,|\,\rpol{t}}=\tgr{t}(\rpol{t},\prc,\arc,\ac)
\ee
\end{lemma}

Note that for the full-information online and batch classification versions of \alg\ we can use the true sub-gradient instead of $\gra{t}$ in order to give a deterministic algorithm. 

Lemmas \ref{sulflem1}, \ref{sulflem2}, \ref{sulflem3} and \ref{sulflem4} combine to give us the following lemma.

\begin{lemma}\label{hdganlem1}
We have:
\be
\mathbb{E}\left[\sum_{\prc\in\prcs}\sum_{\arc\in\cons}\sum_{\ac\in\acs}(\rpol{t}(\prc,\arc,\ac)-\comp(\prc,\arc,\ac))\gra{t}(\prc,\arc,\ac)\,\Bigg|\,\rpol{t}\right]\geq\inr{t}
\ee
\end{lemma}
Now define $\geft{t}:\prcs\times\exs\rightarrow\mathbb{R}$ such that for all $(\prc,\ex)\in\prcs\times\exs$ we have:
\be
\geft{t}(\prc,\ex):=\sum_{\arc\in\cons}\gra{t}(\prc,\arc,\ex(\arc))
\ee

Lemmas \ref{roqslem1} and \ref{hdganlem1} combine to give us the following lemma.
\begin{lemma}\label{hdganlem1.5}
We have:
\be
\mathbb{E}\left[\sum_{\prc\in\prcs}\sum_{\ex\in\exs}(\iwt{t}(\prc,\ex)-\iwfh(\prc,\ex))\geft{t}(\prc,\ex)\Bigg|\,\iwt{t}\right]\geq\inr{t}
\ee
\end{lemma}

The reason that the term $\indi{\cra{t}\leq1}$ appears in the definition of our objective function $\slf{t}$ is to ensure the following crucial property of the pseudo-gradient, which follows from Lemma \ref{nwowrlem1}.

\begin{lemma}\label{hdganlem2}
We have:
\be
\mathbb{E}\left[\sum_{\prc\in\prcs}\sum_{\ex\in\exs}\iwt{t}(\prc,\ex)\geft{t}(\prc,\ex)^2\,\Bigg|\,\iwt{t}\right]\leq 8\nma
\ee
\end{lemma}

Finally we have the following lemma:
\begin{lemma}\label{bbbmklem}
For all $(\prc,\ex)\in\prcs\times\exs$ we have:
\be
\geft{t}(\prc,\ex)\geq-\nma
\ee
\end{lemma}

We have now derived all the properties of $\geft{t}$ needed to progress to the next stage of the algorithm and analysis.

\nc{\arn}{z}

\subsection{Hedge Update}

To update we run, for each $\prc\in\prcs$ and $\arc\in\supp{t}(\prc)$\,, the function: 
\be
\usc{\prc}(\arc,\gra{t}(\prc,\arc,\blank))
\ee
We have the following lemma.
\begin{lemma}\label{rohulem1}
For all $(\prc,\ex)\in\prcs\times\exs$ we have:
\be
\iwt{t+1}(\prc,\ex)=\frac{\iwt{t}(\prc,\ex)\exp(-\lr\geft{t}(\prc,\ex))}{\sum_{\ex'\in\exs}\iwt{t}(\prc,\ex')\exp(-\lr\geft{t}(\prc,\ex'))}
\ee
\end{lemma}

We define the \emp{relative entropy} $\re:\ibs\times\ibs\rightarrow\mathbb{R}$ such that for all $\avt,\avt'\in\ibs$ we have:
\be
\re(\avt,\avt'):=\sum_{\ex\in\exs}\avt(\ex)\ln\left(\frac{\avt(\ex)}{\avt'(\ex)}\right)
\ee
Lemma \ref{bbbmklem} and Lemma \ref{rohulem1} give us the following lemma.
\begin{lemma}\label{fpohblem}
For all $\prc\in\prcs$ we have:
\be
\re(\iwfh,\iwt{t}(\prc,\blank))-\re(\iwfh,\iwt{t+1}(\prc,\blank))\geq\lr\sum_{\ex\in\exs}(\iwt{t}(\prc,\ex)-\iwfh(\prc,\ex))\geft{t}(\prc,\ex)-\lr^2\sum_{\ex\in\exs}\iwt{t}(\prc,\ex)\geft{t}(\prc,\ex)^2
\ee
\end{lemma}

This lemma combines with lemmas \ref{hdganlem1.5} and \ref{hdganlem2} to give the following lemma.
\begin{lemma}\label{flftlem1}
We have:
\be
\mathbb{E}\left[\sum_{\prc\in\prcs}(\re(\iwfh,\iwt{t}(\prc,\blank))-\re(\iwfh,\iwt{t+1}(\prc,\blank)))\,\Bigg|\, \iwt{t}\right]\geq\lr\inr{t}-8\lr^2\nma
\ee
\end{lemma}
This completes the description and analysis of trial $t$.

\subsection{Regret Bound and Fair Classification}

To get the overall regret bound we take expectations on Lemma \ref{flftlem1} and sum over all $t\in[\ntr]$\,, which gives us the following lemma.
\begin{lemma}\label{fb4blem}
We have:
\be
\reg{\comp}\leq(8\pra-\rwcs/\pra)\sqrt{\nma/\ntr}
\ee
where:
\be
\rwcs:=\sum_{\prc\in\prcs}\sum_{\ex\in\exs}\iwfh(\prc,\ex)\ln\left(\frac{\iwfh(\prc,\ex)}{\iwf(\ex)}\right)
\ee
\end{lemma}

Note that in the full-information problem we can maintain $\Theta(\ln(\ntr))$ copies of \alg\ with exponentially increasing learning rates, and combine them together with \textsc{Hedge}. We can also do this for parameters in the inductive bias. Hence, with full-information we have no free parameters. To obtain the fair classification bounds we simply run this full-information version of \alg\ online over the training set with $\edt{t}=\apd$ for all $t\in[\ntr]$, and then output:
\be
\inp:=\frac{1}{\ntr}\sum_{t\in[\ntr]}\polt{t}
\ee
which can be computed online. Note that this is effectively the online-to-batch conversion process given in \cite{CesaBianchi2001OnTG} so the bound  in Theorem \ref{fbclath} comes directly from that and Lemma \ref{fb4blem}.

\nc{\psbtt}[2]{\mathcal{X}'_{#1}(#2)}

\section{Analysis for Approximate Statistical Parity with IID Contexts}\label{afltdwicsec}
We now analyse the quality of the approximation produced by the algorithm given in Section \ref{aspwicssec}. Let $\cdp$ be the height of $\cons$. For any non-root vertex $\arc\in\cons$ let $\pnt{\arc}$ be its parent and for all $\prc\in\prcs$ define: 
\be
\pst(\prc,\arc):=\frac{\trpdc(\prc,\arc)}{\trpdc(\prc,\pnt{\arc})}
\ee
Choose any $\eap>0$. For all $\prc\in\prcs$ let $\psbf{\prc}$ be the subtree $\psbt{\prc}$ on trial $\ntr$. For all $\prc\in\prcs$ and $\arc\in\psbf{\prc}\setminus\{\rot\}$ define:
\be
\den(\prc,\arc):=\empa(\prc,\arc)-\empr(\prc,\arc)
\ee
By Hoeffding's inequality we have, for any $\arc\in\psbf{\prc}\setminus\{\rot\}$, that:
\be
\mathbb{P}[|\den(\prc,\arc)|>\eap]\leq2\exp(-2\eap^2\nsm)
\ee
So since:
\be
\sum_{\prc\in\prcs}|\psbf{\prc}\setminus\{\rot\}|\leq\ntr
\ee
we have, by the union bound, that with probability at least $1-2\ntr\exp(-2\eap^2\nsm)$ we have:
\be
|\den(\prc,\arc)|\leq\eap
\ee
for all $\prc\in\prcs$ and $\arc\in\psbf{\prc}$. So assume that this is the case. Now take any trial $t\in[\ntr]$ and for all $\prc\in\prcs$ let $\psbtt{t}{\prc}$ be the subtree $\psbt{\prc}$ at the start of trial $t$. For all $\prc\in\prcs$ and $\arc\in\psbtt{t}{\prc}$ let $\ndes{t}(\prc,\arc)$ be the set of leaves of $\psbt{\prc}$ that are descendants of $\arc$. For all $\prc\in\prcs$, $\arc\in\psbtt{t}{\prc}$ and $\ac\in\acs$ define:
\be
\ps{t}(\prc,\arc,\ac):=\frac{1}{\trpdc(\prc,\arc)}\sum_{\arc'\in\ndes{t}(\prc,\arc)}\trpdc(\prc,\arc')\polt{t}(\prc,\arc',\ac)
~~~~~;~~~~~\pp{t}(\prc,\arc,\ac):=\frac{1}{\msap(\prc,\arc)}\sum_{\arc'\in\ndes{t}(\prc,\arc)}\msap(\prc,\arc')\polt{t}(\prc,\arc',\ac)
\ee
and define:
\be
\bdf{t}(\prc,\arc,\ac):=\ps{t}(\prc,\arc,\ac)-\pp{t}(\prc,\arc,\ac)
\ee
For all $\prc\in\prcs$ and $\arc\in\psbtt{t}{\prc}$ let $\nvhe{t}(\prc,\arc)$ be the height of $\arc$ in $\psbtt{t}{\prc}$, and take the inductive hypothesis that:
\be
|\bdf{t}(\prc,\arc,\ac)|\leq2\nvhe{t}(\prc,\arc)\eap
\ee
which we now prove by induction over $\nvhe{t}(\prc,\arc)$. The inductive hypothesis clearly holds for $\nvhe{t}(\prc,\arc)=0$ because then $\arc$ is a leaf of $\psbtt{t}{\prc}$ so that $\ndes{t}(\prc,\arc)=\{\arc\}$ and hence:
\be
\ps{t}(\prc,\arc,\ac)=\polt{t}(\prc,\arc,\ac)=\pp{t}(\prc,\arc,\ac)
\ee
Now suppose we have some $\vdp\in\nat\cup\{0\}$ such that the inductive hypothesis holds for $\nvhe{t}(\prc,\arc)=\vdp$. We will now show that it holds for $\nvhe{t}(\prc,\arc)=\vdp+1$ which will complete the proof of the inductive hypothesis. So choose some $\arc\in\psbtt{t}{\prc}$ with $\nvhe{t}(\prc,\arc)=\vdp+1$. Note that:
\begin{align*}
\ps{t}(\prc,\arc,\ac)&=\frac{1}{\trpdc(\prc,\arc)}\left(\trpdc(\prc,\lch{\arc})\ps{t}(\prc,\lch{\arc},\ac)+\trpdc(\prc,\rch{\arc})\ps{t}(\prc,\rch{\arc},\ac)\right)\\
&=\empa(\prc,\lch{\arc})\ps{t}(\prc,\lch{\arc},\ac)+\empa(\prc,\rch{\arc})\ps{t}(\prc,\rch{\arc},\ac)
\end{align*}
and  that:
\begin{align*}
\pp{t}(\prc,\arc,\ac)&=\frac{1}{\msap(\prc,\arc)}\left(\msap(\prc,\lch{\arc})\pp{t}(\prc,\lch{\arc},\ac)+\msap(\prc,\rch{\arc})\pp{t}(\prc,\rch{\arc},\ac)\right)\\
&=\empr(\prc,\lch{\arc})\pp{t}(\prc,\lch{\arc},\ac)+\empr(\prc,\rch{\arc})\pp{t}(\prc,\rch{\arc},\ac)\\
&=\empr(\prc,\lch{\arc})(\ps{t}(\prc,\lch{\arc},\ac)-\bdf{t}(\prc,\lch{\arc},\ac))+\empr(\prc,\rch{\arc})(\ps{t}(\prc,\rch{\arc},\ac)-\bdf{t}(\prc,\rch{\arc},\ac)))
\end{align*}
so, by defining:
\be
\baph(\prc,\arc,\ac):=\empr(\prc,\lch{\arc})\bdf{t}(\prc,\lch{\arc},\ac)+\empr(\prc,\rch{\arc})\bdf{t}(\prc,\rch{\arc},\ac)
\ee
we have:
\begin{align*}
\bdf{t}(\prc,\arc,\ac)&:=\ps{t}(\prc,\arc,\ac)-\pp{t}(\prc,\arc,\ac)\\
&=(\empa(\prc,\lch{\arc})-\empr(\prc,\lch{\arc}))\ps{t}(\prc,\lch{\arc},\ac)+(\empa(\prc,\rch{\arc})-\empr(\prc,\rch{\arc}))\ps{t}(\prc,\rch{\arc},\ac)+\baph(\prc,\arc,\ac)\\
&=\den(\prc,\lch{\arc})\ps{t}(\prc,\lch{\arc},\ac)+\den(\prc,\rch{\arc})\ps{t}(\prc,\rch{\arc},\ac)+\baph(\prc,\arc,\ac)
\end{align*}
so that, by the inductive hypothesis and since $\nvhe{t}(\prc,\lch{\arc})=\nvhe{t}(\prc,\rch{\arc})=\vdp$, we have:
\begin{align*}
|\bdf{t}(\prc,\arc,\ac)|&\leq|\den(\prc,\lch{\arc})|\ps{t}(\prc,\lch{\arc},\ac)+|\den(\prc,\rch{\arc})|\ps{t}(\prc,\rch{\arc},\ac)+|\baph(\prc,\arc,\ac)|\\
&\leq\eap\ps{t}(\prc,\lch{\arc},\ac)+\eap\ps{t}(\prc,\rch{\arc},\ac)+|\baph(\prc,\arc,\ac)|\\
&\leq2\eap+|\baph(\prc,\arc,\ac)|\\
&\leq2\eap+\empr(\prc,\lch{\arc})|\bdf{t}(\prc,\lch{\arc},\ac)|+\empr(\prc,\rch{\arc})|\bdf{t}(\prc,\rch{\arc},\ac)|\\
&\leq2\eap+2\empr(\prc,\lch{\arc})\vdp\eap+2\empr(\prc,\rch{\arc})\vdp\eap\\
&=2(1+\vdp)\eap\\
&=2\nvhe{t}(\prc,\arc)\eap
\end{align*}
We have hence proved that the inductive hypothesis holds always. In particular it holds for $\arc=\rot$ so that:
\begin{equation}\label{bltceeq1}
|\bdf{t}(\prc,\rot,\ac)|\leq2\cdp\eap
\end{equation}

 Letting $\tpolt{t}$ be the actual chosen policy over the set $\tcons$\,, we have:
\begin{align*}
\sum_{\arc\in\cons}\edt{t}(\prc,\arc)\polt{t}(\prc,\arc,\ac)&=\sum_{\arc\in\ndes{t}(\prc,\rot)}\msap(\prc,\arc)\polt{t}(\prc,\arc,\ac)\\
&=\pp{t}(\prc,\rot,\ac)\\
&=\ps{t}(\prc,\rot,\ac)-\bdf{t}(\prc,\rot,\ac)\\
&=\sum_{\arc\in\ndes{t}(\prc,\rot)}\trpdc(\prc,\arc)\polt{t}(\prc,\arc,\ac)-\bdf{t}(\prc,\rot,\ac)\\
&=\sum_{\tarc\in\tcons}\trpdc(\prc,\tarc)\tpolt{t}(\prc,\tarc,\ac)-\bdf{t}(\prc,\rot,\ac)
\end{align*}
So by Equation \eqref{bltceeq1} and since $\polt{t}$ has statistical parity with respect to $\edt{t}$ we have shown that $\tpolt{t}$ has statistical parity with respect to $\trpdc$ up to an additive factor of $4\cdp\eap$.

So we have shown that if we want to achieve statistical parity up to an additive factor of $\eaph$ on every trial with probability at least $(1-\hpv)$ we can choose any $\nsm$ with:
\be
\nsm\geq 8\ln(2\ntr)(\cdp/\eaph)^2
\ee

\section{Proofs}\label{profsec}

We now prove, in order, all of the lemmas given in this paper.

\subsection{Lemma \ref{roqslem1}}
This is immediately clear from the construction of $\rpol{t}$ and the description of \textsc{Hedge} given in Section \ref{habisec}.

\subsection{Lemma \ref{pieflem0}}

We first show that $\imp{t}\in\epl{t}$. To show this consider some $\prc\in\prcs$ and $\ac\in\acs$. We have:
\begin{align*}
(1+\cra{t})\gamn{t}(\imp{t},\prc,\ac)&=(1+\cra{t})\sum_{\arc\in\cons}\edt{t}(\prc,\arc)\imp{t}(\prc,\arc,\ac)\\
&=\sum_{\arc\in\cons}\edt{t}(\prc,\arc)(\rpol{t}(\prc,\arc,\ac)+\cdin{t}(\prc,\ac))\\
&=\sum_{\arc\in\cons}\edt{t}(\prc,\arc)\rpol{t}(\prc,\arc,\ac)+\cdin{t}(\prc,\ac)\sum_{\arc\in\cons}\crf{t}(\prc,\arc)\\
&=\gamn{t}(\rpol{t},\prc,\ac)+\cdin{t}(\prc,\ac)\\
&=\gamn{t}(\rpol{t},\prc,\ac)+\edn{t}(\rpol{t},\prc,\ac)\\
&=\gamn{t}(\rpol{t},\prc,\ac)+\max_{\prc'\in\prcs}\gamn{t}(\rpol{t},\prc',\ac)-\gamn{t}(\rpol{t},\prc,\ac)\\
&=\max_{\prc'\in\prcs}\gamn{t}(\rpol{t},\prc',\ac)
\end{align*}
Since this is independent of $\prc$ we have that $\imp{t}\in\epl{t}$ as required.

We now show that, given $(\prc,\arc)\in\ins$, we have $\sum_{\ac\in\acs}\imp{t}(\prc,\arc,\ac)\leq1$. This is true since $\rpol{t}\in\pols$ and hence:
\begin{align*}
(1+\cra{t})\sum_{\ac\in\acs}\imp{t}(\prc,\arc,\ac)&=\sum_{\ac\in\acs}\rpol{t}(\prc,\arc,\ac)+\sum_{\ac\in\acs}\cdin{t}(\prc,\ac)\\
&=1+\sum_{\ac\in\acs}\edn{t}(\rpol{t},\prc,\ac)\\
&\leq 1+\sum_{\ac\in\acs}\max_{\prc'\in\prcs}\edn{t}(\rpol{t},\prc',\ac)\\
&= 1+ \cer{t}(\rpol{t})\\
&\leq(1+\cra{t})
\end{align*}
as required.

The fact that $\imp{t}(\prc,\arc,\ac)\geq0$ for all $(\prc,\arc,\ac)\in\prcs\times\cons\times\acs$ follows directly from the fact that, since $\rpol{t}\in\pols$, we have that $\cdin{t}(\prc,\ac)$ and $\cra{t}$ and $\rpol{t}(\prc,\arc,\ac)$ are all non-negative.

\subsection{Lemma \ref{pieflem1}}

We first show that $\polt{t}\in\pols$. To show this consider any $(\prc,\arc)\in\ins$. By Lemma \ref{pieflem0} we have that $\sum_{\ac\in\acs}\imp{t}(\prc,\arc,\ac)\leq1$ and hence:
\be
1-\sum_{\ac'\in\acs}\imp{t}(\prc,\arc,\ac')\geq0
\ee
For all $\ac\in\acs$ we have, by Lemma \ref{pieflem0}, that $\imp{t}(\prc,\arc,\ac)\geq0$ and hence the above equation implies that:
\be
\polt{t}(\prc,\arc,\ac)=\imp{t}(\prc,\arc,\ac)+\frac{1}{\nma}\left(1-\sum_{\ac'\in\acs}\imp{t}(\prc,\arc,\ac')\right)\geq 0+0=0
\ee
Finally note that:
\begin{align*}
\notag\sum_{\ac\in\acs}\polt{t}(\prc,\arc,\ac)&=\sum_{\ac\in\acs}\imp{t}(\prc,\arc,\ac)+\frac{1}{\nma}\sum_{\ac\in\acs}\left(1-\sum_{\ac'\in\acs}\imp{t}(\prc,\arc,\ac')\right)\\
&=\sum_{\ac\in\acs}\imp{t}(\prc,\arc,\ac)+\left(1-\sum_{\ac'\in\acs}\imp{t}(\prc,\arc,\ac')\right)\\
&=1
\end{align*}
Since these equations hold for all $(\prc,\arc)\in\ins$, we have shown that $\polt{t}\in\pols$. Hence, all we now need to show is that $\polt{t}\in\epl{t}$. To show this, note that by Lemma \ref{pieflem0} we have that $\imp{t}\in\epl{t}$. Hence, there exists a function $\cdn:\acs\rightarrow\real$ such that for all $(\prc,\ac)\in\prcs\times\acs$ we have:
\be
\sum_{\arc\in\cons}\crf{t}(\prc,\arc)\imp{t}(\prc,\arc,\ac)=\gamn{t}(\imp{t},\prc,\ac)=\cdn(\ac)
\ee
So for all $(\prc,\ac)\in\prcs\times\acs$ we have:
\begin{align*}
\gamn{t}(\polt{t},\prc,\ac)&=\sum_{\arc\in\cons}\crf{t}(\prc,\arc)\polt{t}(\prc,\arc,\ac)\\
&=\sum_{\arc\in\cons}\crf{t}(\prc,\arc)\imp{t}(\prc,\arc,\ac)+\sum_{\arc\in\cons}\frac{\crf{t}(\prc,\arc)}{\nma}\left(1-\sum_{\ac'\in\acs}\imp{t}(\prc,\arc,\ac')\right)\\
&=\cdn(\ac)+\frac{1}{\nma}\sum_{\arc\in\cons}\crf{t}(\prc,\arc)-\frac{1}{\nma}\sum_{\ac'\in\acs}\sum_{\arc\in\cons}\crf{t}(\prc,\arc)\imp{t}(\prc,\arc,\ac')\\
&=\cdn(\ac)+\frac{1}{\nma}-\frac{1}{\nma}\sum_{\ac'\in\acs}\cdn(\ac')
\end{align*}
Since this is independent of $\prc$ we have $\polt{t}\in\epl{t}$ as required. This completes the proof.

\subsection{Lemma \ref{nwowrlem1}}

Choose any $(\prc,\arc,\ac)\in\prcs\times\cons\times\acs$. By Lemma \ref{pieflem0} we have $\sum_{\ac\in\acs}\imp{t}(\prc,\arc,\ac)\leq1$ so since $\cdin{t}(\prc,\ac)$ and $\cra{t}$ and $\rpol{t}(\prc,\arc,\ac)$ are all non-negative, we have that:
\begin{align*}
\polt{t}(\prc,\arc,\ac)&=\imp{t}(\prc,\arc,\ac)+\frac{1}{\nma}\left(1-\sum_{\ac'\in\acs}\imp{t}(\prc,\arc,\ac')\right)\\
&\geq\imp{t}(\prc,\arc,\ac)+\frac{1}{\nma}(1-1)\\
&=\imp{t}(\prc,\arc,\ac)\\
&=\frac{\rpol{t}(\prc,\arc,\ac)+\cdin{t}(\prc,\ac)}{1+\cra{t}}\\
&\geq\frac{\rpol{t}(\prc,\arc,\ac)}{1+\cra{t}}
\end{align*}
as required.

\subsection{Lemma \ref{disclem1}}

Take any $(\prc,\arc,\ac)\in\prcs\times\cons\times\acs$. By Lemma \ref{nwowrlem1} we have:
\be
0\leq\polt{t}(\prc,\arc,\ac)-\frac{\rpol{t}(\prc,\arc,\ac)}{1+\cra{t}}
\ee
Since $\cra{t}$ and $\rpol{t}(\prc,\arc,\ac)$ are non-negative we also have:
\be
\polt{t}(\prc,\arc,\ac)-\rpol{t}(\prc,\arc,\ac)\leq\polt{t}(\prc,\arc,\ac)-\frac{\rpol{t}(\prc,\arc,\ac)}{1+\cra{t}}
\ee
and hence we have shown that:
\be
\max\{0,\polt{t}(\prc,\arc,\ac)-\rpol{t}(\prc,\arc,\ac)\}\leq\polt{t}(\prc,\arc,\ac)-\frac{\rpol{t}(\prc,\arc,\ac)}{1+\cra{t}}
\ee
Now take any $(\prc,\arc)\in\ins$. By lemmas \ref{roqslem1} and \ref{pieflem1} we have that both $\polt{t}$ and $\rpol{t}$ are in $\pols$. Hence, by the above equation we have, since $\cra{t}\geq0$, that:
\begin{align}
\sum_{\ac\in\acs}\max\{0,\polt{t}(\prc,\arc,\ac)-\rpol{t}(\prc,\arc,\ac)\}&\leq\sum_{\ac\in\acs}\left(\polt{t}(\prc,\arc,\ac)-\frac{\rpol{t}(\prc,\arc,\ac)}{1+\cra{t}}\right)\\
&=\sum_{\ac\in\acs}\polt{t}(\prc,\arc,\ac)-\frac{1}{1+\cra{t}}\sum_{\ac\in\acs}\rpol{t}(\prc,\arc,\ac)\\
&=1-\frac{1}{1+\cra{t}}\\
&=\frac{\cra{t}}{1+\cra{t}}\\
&\leq\cra{t}
\end{align}
as required.

\subsection{Lemma \ref{sulflem1}}

We consider two cases. First consider the case that $\cra{t}>1$. In this case we have, since $\rpol{t}\in\pols$ and $\loft{t}(\ac)\leq1$ for all $\ac\in\acs$, that:
\begin{align*}
\slf{t}(\rpol{t})&=\indi{\cra{t}\leq1}\sum_{\ac\in\acs}\loft{t}(\ac)\rpol{t}(\prct{t},\cont{t},\ac)+\cer{t}(\rpol{t})\\
&=\cer{t}(\rpol{t})\\
&=\cra{t}\\
&>1\\
&=\sum_{\ac\in\acs}\polt{t}(\prct{t},\cont{t},\ac)\\
&\geq\sum_{\ac\in\acs}\polt{t}(\prct{t},\cont{t},\ac)\loft{t}(\ac)
\end{align*}

Now consider the case that $\cra{t}\leq1$. Since $\loft{t}(\ac)\in[0,1]$ for all $\ac\in\acs$, we have, by Lemma \ref{disclem1}, that:
\begin{align*}
\sum_{\ac\in\acs}\loft{t}(\ac)\polt{t}(\prct{t},\cont{t},\ac)-\sum_{\ac\in\acs}\loft{t}(\ac)\rpol{t}(\prct{t},\cont{t},\ac)&=\sum_{\ac\in\acs}\loft{t}(\ac)(\polt{t}(\prct{t},\cont{t},\ac)-\rpol{t}(\prct{t},\cont{t},\ac))\\
&\leq\sum_{\ac\in\acs}\loft{t}(\ac)\max\{0,\polt{t}(\prct{t},\cont{t},\ac)-\rpol{t}(\prct{t},\cont{t},\ac)\}\\
&\leq\sum_{\ac\in\acs}\max\{0,\polt{t}(\prct{t},\cont{t},\ac)-\rpol{t}(\prct{t},\cont{t},\ac)\}\\
&\leq\cra{t}
\end{align*}
Hence, we have that:
\begin{align*}
\slf{t}(\rpol{t})&=\indi{\cra{t}\leq1}\sum_{\ac\in\acs}\loft{t}(\ac)\rpol{t}(\prct{t},\cont{t},\ac)+\cer{t}(\rpol{t})\\
&=\sum_{\ac\in\acs}\loft{t}(\ac)\rpol{t}(\prct{t},\cont{t},\ac)+\cra{t}\\
&\geq \sum_{\ac\in\acs}\loft{t}(\ac)\rpol{t}(\prct{t},\cont{t},\ac)+\left(\sum_{\ac\in\acs}\loft{t}(\ac)\polt{t}(\prct{t},\cont{t},\ac)-\sum_{\ac\in\acs}\loft{t}(\ac)\rpol{t}(\prct{t},\cont{t},\ac)\right)\\
&=\sum_{\ac\in\acs}\loft{t}(\ac)\polt{t}(\prct{t},\cont{t},\ac)
\end{align*}
So in either case we have the result.

\subsection{Lemma \ref{sulflem2}}

Since $\comp\in\sps{\spt{t}}$ we have $\cpol{t}\in\epl{t}$ which implies that there exists a function $\cdn:\acs\rightarrow\mathbb{R}$ such that for all $(\prc,\ac)\in\prcs\times\acs$ we have $\gamn{t}(\comp,\prc,\ac)=\cdn(\ac)$. Hence, we have, for all $(\prc,\ac)\in\prcs\times\acs$\,, that:
\be
\edn{t}(\comp,\prc,\ac)=\max_{\prc'\in\prcs}\gamn{t}(\comp,\prc',\ac)-\gamn{t}(\comp,\prc,\ac)=\cdn(\ac)-\cdn(\ac)=0
\ee
Since $\loft{t}(\ac)$ and $\comp(\prct{t},\cont{t},\ac)$ are non-negative for all $\ac\in\acs$, this implies that:
\begin{align*}
\slf{t}(\comp)&:=\indi{\cra{t}\leq1}\sum_{\ac\in\acs}\loft{t}(\ac)\comp(\prct{t},\cont{t},\ac)+\cer{t}(\comp)\\
&\leq\sum_{\ac\in\acs}\loft{t}(\ac)\comp(\prct{t},\cont{t},\ac)+\sum_{\ac\in\acs}\max_{\prc\in\prcs}\edn{t}(\comp,\prc,\ac)\\
&=\sum_{\ac\in\acs}\loft{t}(\ac)\comp(\prct{t},\cont{t},\ac)
\end{align*}
as required.

\subsection{Lemma \ref{sulflem3}}
Take any $\ac\in\acs$. Given $\cpvd\in\eps$ and $\prc'\in\prcs$ we have:
\be
\sum_{(\prc,\arc)\in\ins}\indi{\prc=\prc'}\crf{t}(\prc,\arc)\cpvd(\prc,\arc,\ac)=\sum_{\arc\in\cons}\crf{t}(\prc',\arc)\cpvd(\prc',\arc,\ac)=\gamn{t}(\cpvd,\prc',\ac)
\ee
which implies that:
\begin{equation}\label{sulflem3eq1}
\sum_{(\prc,\arc)\in\ins}\indi{\prc=\kapu{t}(\cpv,\ac)}\crf{t}(\prc,\arc)\cpv(\prc,\arc,\ac)=\max_{\prc'\in\prcs}\gamn{t}(\cpv,\prc',\ac)
\end{equation}
\begin{equation}\label{sulflem3eq2}
\sum_{(\prc,\arc)\in\ins}\indi{\prc=\kapl{t}(\cpv,\ac)}\crf{t}(\prc,\arc)\cpv(\prc,\arc,\ac)=\min_{\prc'\in\prcs}\gamn{t}(\cpv,\prc',\ac)
\end{equation}
\begin{equation}\label{sulflem3eq3}
\sum_{(\prc,\arc)\in\ins}\indi{\prc=\kapu{t}(\cpv,\ac)}\crf{t}(\prc,\arc)\cpv'(\prc,\arc,\ac)\leq\max_{\prc'\in\prcs}\gamn{t}(\cpv',\prc',\ac)
\end{equation}
\begin{equation}\label{sulflem3eq4}
\sum_{(\prc,\arc)\in\ins}\indi{\prc=\kapl{t}(\cpv,\ac)}\crf{t}(\prc,\arc)\cpv'(\prc,\arc,\ac)\geq\min_{\prc'\in\prcs}\gamn{t}(\cpv',\prc',\ac)
\end{equation}
For all $\ac\in\acs$\,, equations \eqref{sulflem3eq1} and \eqref{sulflem3eq2} imply that:
\begin{align*}
&\sum_{(\prc,\arc)\in\ins}\tgr{t}(\cpv,\prc,\arc,\ac)\cpv(\prc,\arc,\ac)\\
&=\sum_{(\prc,\arc)\in\ins}\indi{\cra{t}\leq1}\indi{(\prc,\arc)=(\prct{t},\cont{t})}\loft{t}(\ac)\cpv(\prc,\arc,\ac)+\max_{\prc'\in\prcs}\gamn{t}(\cpv,\prc',\ac)-\min_{\prc'\in\prcs}\gamn{t}(\cpv,\prc',\ac)\\
&=\sum_{(\prc,\arc)\in\ins}\indi{\cra{t}\leq1}\indi{(\prc,\arc)=(\prct{t},\cont{t})}\loft{t}(\ac)\cpv(\prc,\arc,\ac)+\max_{\prc^\dag\in\prcs}\left(\max_{\prc'\in\prcs}\gamn{t}(\cpv,\prc',\ac)-\gamn{t}(\cpv,\prc^\dag,\ac)\right)\\
&=\indi{\cra{t}\leq1}\loft{t}(\ac)\cpv(\prct{t},\cont{t},\ac)+\max_{\prc^\dag\in\prcs}\edn{t}(\cpv,\prc^\dag\ac)
\end{align*}
This means that:
\begin{align}
\notag\sum_{\ac\in\acs}\sum_{(\prc,\arc)\in\ins}\tgr{t}(\cpv,\prc,\arc,\ac)\cpv(\prc,\arc,\ac)&=\sum_{\ac\in\acs}\indi{\cra{t}\leq1}\loft{t}(\ac)\cpv(\prct{t},\cont{t},\ac)+\sum_{\ac\in\acs}\max_{\prc^\dag\in\prcs}\edn{t}(\cpv,\prc^\dag,\ac)\\
\notag&=\indi{\cra{t}\leq1}\sum_{\ac\in\acs}\loft{t}(\ac)\cpv(\prct{t},\cont{t},\ac)+\cer{t}(\cpv)\\
\label{sulflem3eq5}&=\slf{t}(\cpv)
\end{align}
Similarly, for all $\ac\in\acs$, equations \eqref{sulflem3eq3} and \eqref{sulflem3eq4} imply that:
\begin{align*}
&\sum_{(\prc,\arc)\in\ins}\tgr{t}(\cpv,\prc,\arc,\ac)\cpv'(\prc,\arc,\ac)\\
&\leq\sum_{(\prc,\arc)\in\ins}\indi{\cra{t}\leq1}\indi{(\prc,\arc)=(\prct{t},\cont{t})}\loft{t}(\ac)\cpv'(\prc,\arc,\ac)+\max_{\prc'\in\prcs}\gamn{t}(\cpv',\prc',\ac)-\min_{\prc'\in\prcs}\gamn{t}(\cpv',\prc',\ac)\\
&=\sum_{(\prc,\arc)\in\ins}\indi{\cra{t}\leq1}\indi{(\prc,\arc)=(\prct{t},\cont{t})}\loft{t}(\ac)\cpv'(\prc,\arc,\ac)+\max_{\prc^\dag\in\prcs}\left(\max_{\prc'\in\prcs}\gamn{t}(\cpv',\prc',\ac)-\gamn{t}(\cpv',\prc^\dag,\ac)\right)\\
&=\indi{\cra{t}\leq1}\loft{t}(\ac)\cpv'(\prct{t},\cont{t},\ac)+\max_{\prc^\dag\in\prcs}\edn{t}(\cpv,\prc^\dag,\ac)
\end{align*}
This means that:
\begin{align}
\notag\sum_{\ac\in\acs}\sum_{(\prc,\arc)\in\ins}\tgr{t}(\cpv,\prc,\arc,\ac)\cpv'(\prc,\arc,\ac)&\leq\sum_{\ac\in\acs}\indi{\cra{t}\leq1}\loft{t}(\ac)\cpv'(\prct{t},\cont{t},\ac)+\sum_{\ac\in\acs}\max_{\prc^\dag\in\prcs}\edn{t}(\cpv,\prc^\dag,\ac)\\
\notag&=\indi{\cra{t}\leq1}\sum_{\ac\in\acs}\loft{t}(\ac)\cpv'(\prct{t},\cont{t},\ac)+\cer{t}(\cpv')\\
\label{sulflem3eq6}&=\slf{t}(\cpv')
\end{align}
Equations \ref{sulflem3eq5} and \ref{sulflem3eq6} give us:
\begin{align*}
\sum_{(\prc,\arc)\in\ins}\sum_{\ac\in\acs}\tgr{t}(\cpv,\prc,\arc,\ac)(\cpv(\prc,\arc,\ac)-\cpv'(\prc,\arc,\ac))&=\sum_{(\prc,\arc)\in\ins}\sum_{\ac\in\acs}\tgr{t}(\cpv,\prc,\arc,\ac)\cpv(\prc,\arc,\ac)-\sum_{(\prc,\arc)\in\ins}\sum_{\ac\in\acs}\tgr{t}(\cpv,\prc,\arc,\ac)\cpv'(\prc,\arc,\ac)\\
&\geq\slf{t}(\cpv)-\slf{t}(\cpv')
\end{align*}
as required.

\subsection{Lemma \ref{sulflem4}}

In this proof we use, throughout, the property of linearity of expectation. Take any $\ac\in\acs$. Since $\cont{t}$\,, $\prct{t}$\,, $\polt{t}$ and $\lost{t}(\ac)$ are deterministic when conditioned on $\rpol{t}$ we have:
\begin{align}
\notag\bexptc{\indi{\ac=\act{t}}\frac{\lost{t}(\act{t})}{\polt{t}(\prct{t},\cont{t},\act{t})}}{\rpol{t}}&=\bexptc{\indi{\ac=\act{t}}\frac{\lost{t}(\ac)}{\polt{t}(\prct{t},\cont{t},\ac)}}{\rpol{t}}\\
\notag&=\frac{\lost{t}(\ac)}{\polt{t}(\prct{t},\cont{t},\ac)}\exptc{\indi{\ac=\act{t}}}{\rpol{t}}\\
\notag&=\frac{\lost{t}(\ac)}{\polt{t}(\prct{t},\cont{t},\ac)}\probc{\ac=\act{t}}{\rpol{t}}\\
\notag&=\frac{\lost{t}(\ac)}{\polt{t}(\prct{t},\cont{t},\ac)}\polt{t}(\prct{t},\cont{t},\ac)\\
\label{sulflem4eq1}&=\lost{t}(\ac)
\end{align}
Which immediately gives us the result.
\subsection{Lemma \ref{hdganlem1}}

By lemmas \ref{sulflem1} and \ref{sulflem2} we have:
\begin{equation}\label{hdganlem1eq1}
\inr{t}=\sum_{\ac\in\acs}\polt{t}(\prct{t},\cont{t},\ac)\loft{t}(\ac)-\sum_{\ac\in\acs}\comp(\prct{t},\cont{t},\ac)\loft{t}(\ac)\leq\slf{t}(\rpol{t})-\slf{t}(\comp)
\end{equation}
and by lemmas \ref{sulflem3} and \ref{sulflem4} and the linearity of expectation we have, since $\comp$ is independent of $\rpol{t}$, that:
\begin{align}
\notag\slf{t}(\rpol{t})-\slf{t}(\comp)&\leq\sum_{(\prc,\arc)\in\ins}\sum_{\ac\in\acs}\tgr{t}(\rpol{t},\prc,\arc,\ac)(\rpol{t}(\prc,\arc,\ac)-\comp(\prc,\arc,\ac))\\
\notag&=\sum_{(\prc,\arc)\in\ins}\sum_{\ac\in\acs}\expt{\gra{t}(\prc,\arc,\ac)\,|\,\rpol{t}}(\rpol{t}(\prc,\arc,\ac)-\comp(\prc,\arc,\ac))\\
\label{hdganlem1eq2}&=\mathbb{E}\left[\sum_{(\prc,\arc)\in\ins}\sum_{\ac\in\acs}(\rpol{t}(\prc,\arc,\ac)-\comp(\prc,\arc,\ac))\gra{t}(\prc,\arc,\ac)\,\Bigg|\,\rpol{t}\right]
\end{align}
Combining equations \eqref{hdganlem1eq1} and \eqref{hdganlem1eq2} gives us the result.

\subsection{Lemma \ref{hdganlem1.5}}

By Lemma \ref{roqslem1} we have:
\begin{align*}
\sum_{\prc\in\prcs}\sum_{\ex\in\exs}\iwt{t}(\prc,\ex)\geft{t}(\prc,\ex)&=\sum_{\prc\in\prcs}\sum_{\ex\in\exs}\iwt{t}(\prc,\ex)\sum_{\arc\in\cons}\gra{t}(\prc,\arc,\ex(\arc))\\
&=\sum_{\prc\in\prcs}\sum_{\ex\in\exs}\iwt{t}(\prc,\ex)\sum_{\arc\in\cons}\sum_{\ac\in\acs}\indi{\ex(\arc)=\ac}\gra{t}(\prc,\arc,\ac)\\
&=\sum_{(\prc,\arc)\in\ins}\sum_{\ac\in\acs}\gra{t}(\prc,\arc,\ac)\sum_{\ex\in\exs}\iwt{t}(\prc,\ex)\indi{\ex(\arc)=\ac}\\
&=\sum_{(\prc,\arc)\in\ins}\sum_{\ac\in\acs}\gra{t}(\prc,\arc,\ac)\rpol{t}(\prc,\arc,\ac)
\end{align*}
and by definition of $\iwfh$ we have:
\begin{align*}
\sum_{\prc\in\prcs}\sum_{\ex\in\exs}\iwfh(\prc,\ex)\geft{t}(\prc,\ex)&=\sum_{\prc\in\prcs}\sum_{\ex\in\exs}\iwfh(\prc,\ex)\sum_{\arc\in\cons}\gra{t}(\prc,\arc,\ex(\arc))\\
&=\sum_{\prc\in\prcs}\sum_{\ex\in\exs}\iwfh(\prc,\ex)\sum_{\arc\in\cons}\sum_{\ac\in\acs}\indi{\ex(\arc)=\ac}\gra{t}(\prc,\arc,\ac)\\
&=\sum_{(\prc,\arc)\in\ins}\sum_{\ac\in\acs}\gra{t}(\prc,\arc,\ac)\sum_{\ex\in\exs}\iwfh(\prc,\ex)\indi{\ex(\arc)=\ac}\\
&=\sum_{(\prc,\arc)\in\ins}\sum_{\ac\in\acs}\gra{t}(\prc,\arc,\ac)\comp(\prc,\arc,\ac)
\end{align*}
so since $\rpol{t}$ is derived solely from $\iwt{t}$ (and deterministic objects) we have:
\be
\mathbb{E}\left[\sum_{\prc\in\prcs}\sum_{\ex\in\exs}(\iwt{t}(\prc,\ex)-\iwfh(\prc,\ex))\geft{t}(\prc,\ex)\Bigg|\,\iwt{t}\right]=\mathbb{E}\left[\sum_{\prc\in\prcs}\sum_{\arc\in\cons}\sum_{\ac\in\acs}(\rpol{t}(\prc,\arc,\ac)-\comp(\prc,\arc,\ac)\gra{t}(\prc,\arc,\ac)\,\Bigg|\,\rpol{t}\right]
\ee
so Lemma \ref{hdganlem1} gives us the result.

\subsection{Lemma \ref{hdganlem2}}

Since $(z+\hat{z})^2\leq 2z^2+2\hat{z}^2$ for all $z,\hat{z}\in\mathbb{R}$ and since $(z-\hat{z})^2\leq z^2 + \hat{z}^2$ for all $z,\hat{z}>0$, we have:
\begin{align}
\notag\sum_{\prc\in\prcs}\sum_{\ex\in\exs}\iwt{t}(\prc,\ex)\geft{t}(\prc,\ex)^2&=\sum_{\prc\in\prcs}\sum_{\ex\in\exs}\iwt{t}(\prc,\ex)\left(\sum_{\arc\in\cons}\gra{t}(\prc,\arc,\ex(\arc))\right)^2\\
\notag&=\sum_{\prc\in\prcs}\sum_{\ex\in\exs}\iwt{t}(\prc,\ex)\left(\sum_{\arc\in\cons}\sum_{\ac\in\acs}\indi{\ex(\arc)=\ac}\gra{t}(\prc,\arc,\ac)\right)^2\\
\notag&\leq2\sum_{\prc\in\prcs}\sum_{\ex\in\exs}\iwt{t}(\prc,\ex)\left(\sum_{\arc\in\cons}\sum_{\ac\in\acs}\indi{\ex(\arc)=\ac}\indi{\cra{t}\leq1}\indi{(\prc,\arc,\ac)=(\prct{t},\cont{t},\act{t})}\frac{\loft{t}(\act{t})}{\polt{t}(\prct{t},\cont{t},\act{t})}\right)^2\\
\notag&~~~~~~+2\sum_{\prc\in\prcs}\sum_{\ex\in\exs}\iwt{t}(\prc,\ex)\left(\sum_{\arc\in\cons}\sum_{\ac\in\acs}\indi{\ex(\arc)=\ac}\indi{\prc=\kapu{t}(\rpol{t},\ac)}\edt{t}(\prc,\arc)\right)^2\\
\label{cwtspeq1}&~~~~~~+2\sum_{\prc\in\prcs}\sum_{\ex\in\exs}\iwt{t}(\prc,\ex)\left(\sum_{\arc\in\cons}\sum_{\ac\in\acs}\indi{\ex(\arc)=\ac}\indi{\prc=\kapl{t}(\rpol{t},\ac)}\edt{t}(\prc,\arc)\right)^2
\end{align}
Note that:
\begin{align}
\notag&\sum_{\prc\in\prcs}\sum_{\ex\in\exs}\iwt{t}(\prc,\ex)\left(\sum_{\arc\in\cons}\sum_{\ac\in\acs}\indi{\ex(\arc)=\ac}\indi{\prc=\kapu{t}(\rpol{t},\ac)}\edt{t}(\prc,\arc)\right)^2\\
\notag&=\sum_{\prc\in\prcs}\sum_{\ex\in\exs}\iwt{t}(\prc,\ex)\left(\sum_{\arc\in\cons}\sum_{\ac\in\acs}\indi{\ex(\arc)=\ac}\indi{\prc=\kapu{t}(\rpol{t},\ac)}\edt{t}(\prc,\arc)\right)\left(\sum_{\arc\in\cons}\sum_{\ac\in\acs}\indi{\ex(\arc)=\ac}\indi{\prc=\kapu{t}(\rpol{t},\ac)}\edt{t}(\prc,\arc)\right)\\
\notag&\leq\sum_{\prc\in\prcs}\sum_{\ex\in\exs}\iwt{t}(\prc,\ex)\left(\sum_{\arc\in\cons}\sum_{\ac\in\acs}\indi{\prc=\kapu{t}(\rpol{t},\ac)}\edt{t}(\prc,\arc)\right)\left(\sum_{\arc\in\cons}\edt{t}(\prc,\arc)\sum_{\ac\in\acs}\indi{\ex(\arc)=\ac}\right)\\
\notag&=\sum_{\prc\in\prcs}\sum_{\ex\in\exs}\iwt{t}(\prc,\ex)\left(\sum_{\ac\in\acs}\indi{\prc=\kapu{t}(\rpol{t},\ac)}\sum_{\arc\in\cons}\edt{t}(\prc,\arc)\right)\left(\sum_{\arc\in\cons}\edt{t}(\prc,\arc)\right)\\
\notag&=\sum_{\prc\in\prcs}\sum_{\ex\in\exs}\iwt{t}(\prc,\ex)\sum_{\ac\in\acs}\indi{\prc=\kapu{t}(\rpol{t},\ac)}\\
\notag&=\sum_{\ac\in\acs}\sum_{\prc\in\prcs}\indi{\prc=\kapu{t}(\rpol{t},\ac)}\sum_{\ex\in\exs}\iwt{t}(\prc,\ex)\\
\notag&=\sum_{\ac\in\acs}\sum_{\prc\in\prcs}\indi{\prc=\kapu{t}(\rpol{t},\ac)}\\
\notag&=\sum_{\ac\in\acs}1\\
\label{cwtspeq2}&=\nma
\end{align}
Similarly we have:
\begin{equation}\label{cwtspeq3}
\sum_{\prc\in\prcs}\sum_{\ex\in\exs}\iwt{t}(\prc,\ex)\left(\sum_{\arc\in\cons}\sum_{\ac\in\acs}\indi{\ex(\arc)=\ac}\indi{\prc=\kapl{t}(\rpol{t},\ac)}\edt{t}(\prc,\arc)\right)^2=\nma
\end{equation}
Now note that, by Lemma \ref{nwowrlem1}, we have, for all $\ac\in\acs$ that:
\be
\frac{\indi{\cra{t}\leq1}}{\polt{t}(\prct{t},\cont{t},\ac)}\leq\frac{2}{\rpol{t}(\prct{t},\cont{t},\ac)}\
\ee
so by Lemma \ref{roqslem1} we have:
\begin{align}
\notag&\mathbb{E}\left[\sum_{\prc\in\prcs}\sum_{\ex\in\exs}\iwt{t}(\prc,\ex)\left(\sum_{\arc\in\cons}\sum_{\ac\in\acs}\indi{\ex(\arc)=\ac}\indi{\cra{t}\leq1}\indi{(\prc,\arc,\ac)=(\prct{t},\cont{t},\act{t})}\frac{\loft{t}(\act{t})}{\polt{t}(\prct{t},\cont{t},\act{t})}\right)^2\,\Bigg|\,\iwt{t}\right]\\
\notag&=\mathbb{E}\left[\sum_{\ex\in\exs}\iwt{t}(\prct{t},\ex)\left(\indi{\ex(\cont{t})=\act{t}}\indi{\cra{t}\leq1}\frac{\loft{t}(\act{t})}{\polt{t}(\prct{t},\cont{t},\act{t})}\right)^2\,\Bigg|\,\iwt{t}\right]\\
\notag&=\sum_{\ac\in\acs}\mathbb{P}[\act{t}=\ac\,|\,\iwt{t}]\sum_{\ex\in\exs}\iwt{t}(\prct{t},\ex)\left(\indi{\ex(\cont{t})=\ac}\indi{\cra{t}\leq1}\frac{\loft{t}(\ac)}{\polt{t}(\prct{t},\cont{t},\ac)}\right)^2\\
\notag&=\sum_{\ac\in\acs}\polt{t}(\prct{t},\cont{t},\ac)\sum_{\ex\in\exs}\iwt{t}(\prct{t},\ex)\left(\indi{\ex(\cont{t})=\ac}\indi{\cra{t}\leq1}\frac{\loft{t}(\ac)}{\polt{t}(\prct{t},\cont{t},\ac)}\right)^2\\
\notag&=\sum_{\ac\in\acs}\sum_{\ex\in\exs}\iwt{t}(\prct{t},\ex)\indi{\ex(\cont{t})=\ac}\indi{\cra{t}\leq1}\frac{\loft{t}(\ac)^2}{\polt{t}(\prct{t},\cont{t},\ac)}\\
\notag&\leq\sum_{\ac\in\acs}\sum_{\ex\in\exs}\iwt{t}(\prct{t},\ex)\indi{\ex(\cont{t})=\ac}\frac{\indi{\cra{t}\leq1}}{\polt{t}(\prct{t},\cont{t},\ac)}\\
\notag&\leq\sum_{\ac\in\acs}\frac{2}{\rpol{t}(\prct{t},\cont{t},\ac)}\sum_{\ex\in\exs}\iwt{t}(\prct{t},\ex)\indi{\ex(\cont{t})=\ac}\\
\notag&\leq\sum_{\ac\in\acs}\frac{2}{\rpol{t}(\prct{t},\cont{t},\ac)}\rpol{t}(\prct{t},\cont{t},\ac)\\
\notag&\leq\sum_{\ac\in\acs}2\\
\label{cwtspeq4}&=2\nma
\end{align}
Taking expectations on Equation \eqref{cwtspeq1} and then substituting in equations \eqref{cwtspeq2}, \eqref{cwtspeq3} and \eqref{cwtspeq4} gives us the result.

\subsection{Lemma \ref{bbbmklem}}

Take any $(\prc,\ex)\in\prcs\times\exs$. Note first that for all $(\arc,\ac)\in\cons\times\acs$ we have:
\be
\gra{t}(\prc,\arc,\ac)\geq-\indi{\prc=\kapl{t}(\rpol{t},\ac)}\edt{t}(\prc,\arc)\geq-\edt{t}(\prc,\arc)
\ee
so that:
\begin{align*}
\geft{t}(\prc,\ex)&=\sum_{\arc\in\cons}\gra{t}(\prc,\arc,\ex(\arc))\\
&\geq\sum_{\ac\in\acs}\sum_{\arc\in\cons}\gra{t}(\prc,\arc,\ac)\\
&\geq-\sum_{\ac\in\acs}\sum_{\arc\in\cons}\edt{t}(\prc,\arc)\\
&=-\sum_{\ac\in\acs}1\\
&\geq-\nma
\end{align*}
as required.

\subsection{Lemma \ref{rohulem1}}

By the description of the \ups\ subroutine in Section \ref{habisec} we see that our update procedure is equivalent to creating a function $\iwtp{t}\in\mathbb{R}^{\prcs\times\exs}$ defined such that for all $(\prc,\ex)\in\prcs\times\exs$ we have:
\be
\iwtp{t}(\prc,\ex)=\iwt{t}(\prc,\ex)\prod_{\arc\in\supp{t}(\prc)}\exp(-\lr\gra{t}(\prc,\arc,\ex(\arc)))
\ee
and then, for all $\prc\in\prcs$\,, normalising $\iwtp{t}(\prc,\blank)$ to create $\iwt{t}(\prc,\blank)$. Note that for all $(\prc,\arc)\in\ins$ with $\arc\notin\supp{t}(\prc)$ we have $\edt{t}(\prc,\arc)=0$ and $(\prc,\arc)\neq(\prct{t},\cont{t})$ so that $\gra{t}(\prc,\arc,\ac)=0$ for all $\ac\in\acs$. This implies that for all $(\prc,\ex)\in\prcs\times\exs$ we have:
\be
\iwtp{t}(\prc,\ex)=\iwt{t}(\prc,\ex)\prod_{\arc\in\cons}\exp(-\lr\gra{t}(\prc,\arc,\ex(\arc)))=\iwt{t}(\prc,\ex)\exp\left(-\lr\sum_{\arc\in\cons}\gra{t}(\prc,\arc,\ex(\arc))\right)=\iwt{t}(\prc,\ex)\exp(-\lr\geft{t}(\prc,\ex))
\ee
so that:
\be
\iwt{t+1}(\prc,\ex)=\frac{\iwt{t}(\prc,\ex)\exp(-\lr\geft{t}(\prc,\ex))}{\sum_{\ex'\in\exs}\iwt{t}(\prc,\ex')\exp(-\lr\geft{t}(\prc,\ex'))}
\ee
as required.

\subsection{Lemma \ref{fpohblem}}

Take any $\prc\in\prcs$. Define:
\be
\arn :=  \sum_{\ex\in\exs}\iwt{t}(\prc,\ex)\exp(-\lr\geft{t}(\prc,\ex))
\ee
By Lemma \ref{rohulem1} we have:
\begin{align}
\notag\re(\iwfh,\iwt{t}(\prc,\blank))-\re(\iwfh,\iwt{t+1}(\prc,\blank))&=\sum_{\ex\in\exs}\iwfh(\prc,\ex)\left(\ln\left(\frac{\iwfh(\prc,\ex)}{\iwt{t}(\prc,\ex)}\right)-\ln\left(\frac{\iwfh(\prc,\ex)}{\iwt{t+1}(\prc,\ex)}\right)\right)\\
\notag&=\sum_{\ex\in\exs}\iwfh(\prc,\ex)\ln\left(\frac{\iwt{t+1}(\prc,\ex)}{\iwt{t}(\prc,\ex)}\right)\\
\notag&=\sum_{\ex\in\exs}\iwfh(\prc,\ex)\ln\left(\frac{\exp(-\lr\geft{t}(\prc,\ex))}{\arn}\right)\\
\notag&=-\lr\sum_{\ex\in\exs}\iwfh(\prc,\ex)\geft{t}(\prc,\ex)-\ln(\arn)\sum_{\ex\in\exs}\iwfh(\prc,\ex)\\
\label{fpohdeq1}&=-\lr\sum_{\ex\in\exs}\iwfh(\prc,\ex)\geft{t}(\prc,\ex)-\ln(\arn)
\end{align}
Since $\pra\leq\sqrt{\ntr/\nma}$ we have $\lr\leq1/\nma$ so that, by Lemma \ref{bbbmklem}, we have, for all $\ex\in\exs$ that:
\be
\lr\geft{t}(\prc,\ex)\geq-1
\ee
so, since $\exp(-\hat{\arn})\leq 1-\hat{\arn}+\hat{\arn}^2$ for all $\arc\geq-1$ and $\ln(1+\hat{\arn
})\leq\hat{\arn}$ for all $\hat{\arn}\in\mathbb{R}$\,, we have:
\begin{align}
\notag\ln(\arn)&=\ln\left(\sum_{\ex\in\exs}\iwt{t}(\prc,\ex)\exp(-\lr\geft{t}(\prc,\ex))\right)\\
\notag&\leq\ln\left(\sum_{\ex\in\exs}\iwt{t}(\prc,\ex)(1-\lr\geft{t}(\prc,\ex)+\lr^2\geft{t}(\prc,\ex)^2)\right)\\
\notag&=\ln\left(\sum_{\ex\in\exs}\iwt{t}(\prc,\ex)+ \sum_{\ex\in\exs}\iwt{t}(\prc,\ex)(-\lr\geft{t}(\prc,\ex)+\lr^2\geft{t}(\prc,\ex)^2)\right)\\
\notag&=\ln\left(1+\sum_{\ex\in\exs}\iwt{t}(\prc,\ex)(-\lr\geft{t}(\prc,\ex)+\lr^2\geft{t}(\prc,\ex)^2)\right)\\
\notag&\leq\sum_{\ex\in\exs}\iwt{t}(\prc,\ex)(-\lr\geft{t}(\prc,\ex)+\lr^2\geft{t}(\prc,\ex)^2)\\
\label{fpohdeq2}&=-\lr\sum_{\ex\in\exs}\iwt{t}(\prc,\ex)\geft{t}(\prc,\ex)+\lr^2\sum_{\ex\in\exs}\iwt{t}(\prc,\ex)\geft{t}(\prc,\ex)^2
\end{align}
Substituting Equation \eqref{fpohdeq2} into Equation \eqref{fpohdeq1} gives us the result.

\subsection{Lemma \ref{flftlem1}}

Immediate from summing the inequality in Lemma \ref{fpohblem} over all $\prc\in\prcs$, taking expectations (conditioned on $\iwt{t}$) and substituting in the inequalities in lemmas \ref{hdganlem1.5} and \ref{hdganlem2}.

\subsection{Lemma \ref{fb4blem}}

From Lemma \ref{flftlem1} we have, for all $t\in[\ntr]$, that:
\be
\lr\mathbb{E}[\inr{t}]-8\lr^2\nma=\mathbb{E}\left[\sum_{\prc\in\prcs}(\re(\iwfh,\iwt{t}(\prc,\blank))-\re(\iwfh,\iwt{t+1}(\prc,\blank)))\right]
\ee
and hence, by summing over all $t\in[\ntr]$ and noting the linearity of expectation and the fact that the relative entropy is non-negative, we have:
\begin{align*}
\lr\reg{\comp}-8\lr^2\nma\ntr&=\mathbb{E}\left[\sum_{\prc\in\prcs}\sum_{t\in[\ntr]}(\re(\iwfh,\iwt{t}(\prc,\blank))-\re(\iwfh,\iwt{t+1}(\prc,\blank)))\right]\\
&=\mathbb{E}\left[\sum_{\prc\in\prcs}(\re(\iwfh,\iwt{1}(\prc,\blank))-\re(\iwfh,\iwt{\ntr+1}(\prc,\blank)))\right]\\
&\leq\mathbb{E}\left[\sum_{\prc\in\prcs}\re(\iwfh,\iwt{1}(\prc,\blank))\right]\\
&=\sum_{\prc\in\prcs}\re(\iwfh,\iwf)\\
\end{align*}
so that:
\be
\reg{\comp}\leq\frac{1}{\lr}\sum_{\prc\in\prcs}\re(\iwfh,\iwf)+8\lr\nma\ntr
\ee
noting that $\lr=\pra/\sqrt{\nma\ntr}$, we have the result.

\end{document}